\documentclass[preprint,12pt,3p]{elsarticle}

\usepackage{lineno,hyperref}
\modulolinenumbers[5]

\usepackage{times}
\usepackage{hyperref}
\usepackage{amssymb, amsmath, amsbsy}
\usepackage{makecell}  
\usepackage{graphicx}
\usepackage{float}
\usepackage{multirow}
\usepackage{xcolor}
\usepackage{soul}
\usepackage{algorithm}
\usepackage{algpseudocode}
\usepackage{natbib}
\usepackage{cleveref}
\usepackage{tikz}

\usepackage{amsmath}

\DeclareMathOperator*{\argmin}{arg\,min}

\newcommand{\bs}[1]{\boldsymbol{#1}}
\newcommand{\mc}[1]{\mathcal{#1}}

\journal{ASCE}

\begin{document}

\begin{frontmatter}

\title{Physics-informed neural network solution of thermo-hydro-mechanical (THM) processes in porous media}

\date{March 1, 2022}

\author[label5]{\href{https://orcid.org/0000-0002-4352-5435}{\includegraphics[scale=0.06]{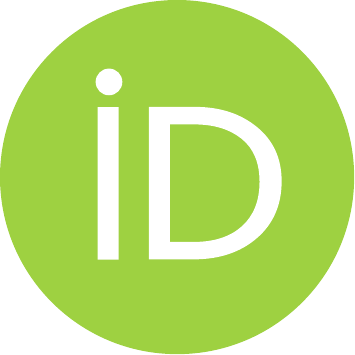}\hspace{1mm}Danial Amini}}
\address[label5]{Department of Civil and Environmental Engineering, Sharif University of Technology, Tehran, Iran}
\ead{aminidaniel.civil@gmail.com}

\author[label2]{\href{https://orcid.org/0000-0003-2659-0507}{\includegraphics[scale=0.06]{orcid.pdf}\hspace{1mm}Ehsan Haghighat}
\corref{cor1}}
\address[label2]{Department of Civil and Environmental Engineering, Massachusetts Institute of Technology, Cambridge MA, USA}
\ead{ehsanh@mit.edu}

\author[label2]{
\href{https://orcid.org/0000-0002-7370-2332}{\includegraphics[scale=0.06]{orcid.pdf}\hspace{1mm}Ruben Juanes}}
\ead{juanes@mit.edu}

\cortext[cor1]{Corresponding to: Ehsan Haghighat  (ehsanh@mit.edu)}

\begin{abstract}
Physics-Informed Neural Networks (PINNs) have received increased interest for forward, inverse, and surrogate modeling of problems described by partial differential equations (PDE). However, their application to multiphysics problem, governed by several \emph{coupled} PDEs, present unique challenges that have hindered the robustness and widespread applicability of this approach. Here we investigate the application of PINNs to the forward solution of problems involving thermo-hydro-mechanical (THM) processes in porous media, which exhibit disparate spatial and temporal scales in thermal conductivity, hydraulic permeability, and elasticity. In addition, PINNs are faced with the challenges of the multi-objective and non-convex nature of the optimization problem. To address these fundamental issues, we: (1)~rewrite the THM governing equations in dimensionless form that is best suited for deep-learning algorithms; (2)~propose a sequential training strategy that circumvents the need for a simultaneous solution of the multiphysics problem and facilitates the task of optimizers in the solution search; and (3)~leverage adaptive weight strategies to overcome the stiffness in the gradient flow of the multi-objective optimization problem. Finally, we apply this framework to the solution of several synthetic problems in 1D and~2D.
\end{abstract}


\end{frontmatter}

\linenumbers


\section{Introduction}

Coupled thermo-hydro-mechanical (THM) processes in porous media arise in a wide range of subsurface applications, including reservoir engineering, geothermal engineering, and deep nuclear waste disposal systems, to name a few. In reservoir engineering, such interactions control the state of subsurface fluids and gases, the performance of production wells, and surface deformation \cite{wang2009parallel, gelet2012thermo}. In geothermal systems, the underground heat is transported to the surface through fluid flow, and optimization of such systems requires a comprehensive THM analysis \cite{o2001state,pandey2017coupled, praditia2018multiscale}. Deep geologic formations have also been proposed as potential sites for nuclear waste disposal, where their long-term performance and safety require thorough THM considerations \cite{rutqvist2009comparative, ballarini2017thermal}.

The mathematical formulation of THM processes results in a system of coupled nonlinear partial differential equations (PDEs) that govern the interaction between different components \cite{rutqvist2001thermohydromechanics}. Classically, such problems are solved by applying a spatial discretization scheme such as the finite element or the finite volume methods, a time-integration scheme such as the implicit or explicit Euler methods, and a linearization scheme such as the Newton's method \cite{Lewis-book}. The resulting coupled system of algebraic equation is often ill-conditioned, a challenge that has sparked the development of iterative sequential solvers, whose stability and convergence depend on the details of the operator splitting and the coupling strength of the multiphysics problem \cite{Lewis-book,15-THM-2018,23-THM,kim2018unconditionally}.

A recent trend in computational science explores the application of machine learning methods for the solution of PDEs. Among the different approaches (e.g., \cite{yu2017deep, ranade2021discretizationnet}), Physics-Informed Neural Networks (PINNs), introduced by \citet{IntroductionPINN}, provide a unified framework for the solution and inversion of boundary value problems. A PINN solver uses fully connected neural networks to approximate the solution variables, and is trained using the strong form of the governing equations, which are evaluated readily using Automatic Differentiation \cite{baydin2018automatic}. This convenience has made the framework very popular and has been explored extensively for a wide range of problems, including fluid mechanics (e.g., \cite{jin2021nsfnets, wu2018physics, cai2021physics, rao2020physics, mao2020physics, reyes2021learning, eivazi2021physics}), solid mechanics (e.g., \cite{haghighat2021physics, rao2021physics}), and heat transfer \cite{cai2021physicsheat, niaki2021physics} (for a detailed review see \cite{39-PINN-TotalReview}). 

As forward solvers, PINNs exhibit limitations for problems developing sharp gradients, or problems involving coupled PDEs. Therefore,  the original version of PINNs has been extended using domain decomposition approaches \cite{33-PINN-XPINN} or nonlocal formulations \cite{haghighat2021nonlocal} to improve their accuracy. Another remedy has involved the use of adaptive weights on the optimization problem \cite{1-PINN-AddaptiveActivationFunction} or new network architectures with feature imbedding \cite{10-PINN-FourierNetwork}. For coupled problems, sequential training has shown improved performance in the robustness and accuracy of PINNs \cite{ 23-PINN-mulit-species-flow-heat-sequential, our-HM-paper}. 

The application of PINNs to poromechanics remains very limited. \citet{41-PINNTerzaghi-forward-inverse} evaluated the performance of a PINN solver on the Terzaghi consolidation problem. \citet{15-PINN-Hamdi-limitations} simulated two-phase fluid flow with various flux functions that result in shocks and refractions. In a set of follow-up studies, they also explored inversion of flow parameters without and with noisy data \cite{16-PINN-Hamdi-2D-2phases,14-PINN-Hamdi-GAN-inversion,43-PINN-similar-Hamdi}. \citet{24-PINN-Biot-forward-inversion} considered single-phase fluid flow in deformable porous media through the lens of Biot’s equations. \citet{36-PINN-BarryMercer} applied PINNs for inversion of flow and deformation data assuming incompressibility of both fluid and solid skeleton. Most recently, we presented a sequential training strategy for the solution of multiphase poromechanics using PINN solvers \cite{our-HM-paper}. None of these studies explore thermo-hydro-mechanical coupling, and this is therefore our focus here.

In this study, our main objective is solving the coupled equations of heat transfer, fluid flow, and solid deformation in porous media using PINNs. Here, we extend our previous work on PINN solvers for multiphase flow in deformable porous media  \cite{our-HM-paper} to also account for thermal coupling. Therefore, the sequential fixed-stress-split strategy, which has been proposed in the context of traditional PDE solvers \citep{kim-fixedStress}, is also employed here to resolve the challenges associated with training PINNs. We use the dimensionless form of the governing equations, which include the energy equation, multiphase fluid flow relations, and the linear Navier relation for the elastic deformation of the solid phase. We apply the proposed framework to the solution of several benchmark problems and show that, in the PINN framework, the sequential training leads to improved robustness and accuracy compared with the simultaneous-solution formulation \cite{our-HM-paper}.

\section{Non-Isothermal Two-Phase Flow in Porous Media}
In this section, we provide the mathematical description of non-isothermal two-phase flow in poroelastic media and its dimensionless forms, best suited for machine learning algorithms. 

\subsection{Balance Laws}

The governing equations describing the Thermo-Hydro-Mechanical response of a porous medium under non-isothermal conditions include linear and angular momentum balance, mass conservation, and energy balance laws. Under the assumption of quasi-static evolution, the linear momentum balance equation is written as
\begin{align}
{\boldsymbol{\nabla}}{\cdot}{\boldsymbol{\sigma}} + {\rho_b}{g}{\boldsymbol{d}} = {\mathbf{0}}, \label{eq1}
\end{align}
in which ${\boldsymbol{\sigma}}$ denotes the total Cauchy stress tensor, ${\rho_b}$ is the bulk density, and ${g}$ represents the gravity acceleration in the direction ${\boldsymbol{d}}$. To satisfy the angular momentum balance relation, the Cauchy stress tensor must be symmetric, i.e., ${\boldsymbol{\sigma^T} = \boldsymbol{\sigma}}$. 

Assuming that the fluids are immiscible, the mass conservation law for each phase~$\alpha$ takes the form \cite{jhajuanes14-wrr}
\begin{align}
{\frac{d m_{\alpha}}{d t}} + {\boldsymbol{\nabla}}{\cdot}{\boldsymbol{w}_{\alpha}} = {\rho_{\alpha}}{f_{\alpha}}, \label{eq2}
\end{align}
where ${m_{\alpha}}$ is the mass of fluid phase $\alpha$ per unit bulk volume, ${\boldsymbol{w}_{\alpha}}$ is the fluid mass flux of phase $\alpha$, and ${f_{\alpha}}$ is a volumetric source term for phase $\alpha$.

Lastly, considering ${m_\theta}$ as energy per unit bulk volume and ${\boldsymbol{h}_{\theta}}$ as the heat flux, the energy balance equation is expressed as 
\begin{align}
{\frac{d {m_\theta}}{d t}} + {\boldsymbol{\nabla}}{\cdot}{\boldsymbol{h}_{\theta}} = {G_\theta}, \label{eq3}
\end{align}
in which ${G_{\theta}}$ represents the volumetric heat source.

\subsection{Small Deformation Kinematics}

Considering the matrix displacement field ${\boldsymbol{u}}$ and under the assumption of small deformations, the matrix strain tensor ${\boldsymbol{\varepsilon}}$ can be written as 
\begin{align}
& {\boldsymbol{\varepsilon}} = {({\boldsymbol{\nabla}}{\boldsymbol{u}}+{\boldsymbol{\nabla}}{\boldsymbol{u}}^{T})}/{2}, \label{eq4}
\end{align}
which can be decomposed into a volumetric part  $\varepsilon_v$ and a deviatoric part  $\boldsymbol{e}$, expressed as 
\begin{align}
& {\varepsilon_v} = {\text{tr}({\boldsymbol{\varepsilon}})}, \label{eq5}\\
& {\boldsymbol{e}} = {\boldsymbol{\varepsilon}} - \frac{\varepsilon_v}{3}{\mathbf{1}}. \label{eq6}
\end{align}

\subsection{Poroelastic Constitutive Relations}

When multiple fluid phases occupy the pore space, there are two approaches to define the pressure: the saturation-averaged pore pressure and the equivalent pore pressure \cite{coussy2004poromechanics}. The accuracy, convergence, and stability of each definition has been studied in \cite{kim2013rigorous}. Here, we use the equivalent pore pressure concept, defined as \cite{coussy2004poromechanics}
\begin{align} 
{p_E} = {\sum_{\alpha}{S_{\alpha}}{p_{\alpha}}}-{U}, \label{eq7}
\end{align}
in which ${S_{\alpha}}$ and ${p_{\alpha}}$ are saturation and fluid pressure of phase $\alpha$, respectively. The interfacial energy ${U}$ is expressed, in incremental form, as ${\delta U = \sum_{\alpha}{p_{\alpha}}{\delta {S_{\alpha}}}}$. For two-phase flow in porous media, the wetting phase and nonwetting phases are denoted by $w$ and $n$, respectively, and therefore the capillary pressure $p_c$ is defined as the pressure difference of nonwetting and wetting fluids, as 
\begin{align}
{p_c} = {p_n} - {p_w}.\label{eq8}
\end{align}
The poroelastic constitutive relation for a two-phase system, considering thermal stresses, is expressed as 
\begin{align}
{\delta\boldsymbol{\sigma}}= {K_{dr} \varepsilon_v \mathbf{1}} + {3 \nu^* K_{dr}}{\boldsymbol{e}} - b\delta p_E  \mathbf{1} - {\beta_s}{K_{dr}}{\delta T \mathbf{1}}, \label{eq9}
\end{align}
or equivalently,
\begin{align}
{\boldsymbol{\sigma}-\boldsymbol{\sigma_0}} = {K_{dr} \varepsilon_v \mathbf{1}} + {3 \nu^* K_{dr}}{\boldsymbol{e}} - {b}\sum_{\alpha}{S_{\alpha} (p_{\alpha} - {p_{\alpha}}_0)} {\mathbf{1}} - {\beta_s K_{dr} {(T - T_0)}}{\mathbf{1}}, \label{eq10}
\end{align}
where $b\delta p_E= \sum_{\alpha} b_{\alpha} \delta p_{\alpha}$, in which ${b}$ is the Biot's coefficient, $K_{dr}$ is the drained bulk modulus, and ${\beta_s}$ denotes the thermal expansion coefficient of the solid phase. 
Parameter $\nu^*$ is defined as $\nu^* = (1-2\nu)/(1+\nu)$ with $\nu$ as Poisson's ratio, and therefore, we have the shear modulus as $G=3/2~\nu^* K_{dr}$.
The volumetric part of the stress tensor \eqref{eq9} is expressed as 
\begin{align}
{\delta \sigma_v} = {K_{dr}}{\varepsilon_v} - {b}\delta p_E - {\beta_s}{K_{dr}}{\delta T}.  \label{eq11}
\end{align}

\subsection{Two-Phase Flow Constitutive Relations}
The mass flux in \cref{eq2} is expressed as ${\boldsymbol{w}_{\alpha}} = {\rho_{\alpha}}{\boldsymbol{v}_{\alpha}}$, in which $\boldsymbol{v}_{\alpha}$ is the Darcy velocity given as 
\begin{align}
{\boldsymbol{v}_{\alpha}} = -\frac{k}{\mu_{\alpha}}{k_{\alpha}^r}{({\boldsymbol{\nabla}{p_{\alpha}}}-{\rho_{\alpha}}{g}{\boldsymbol{d}})}, \label{eq12}
\end{align}
where ${k}$ denotes the intrinsic permeability of the porous medium, $k_{\alpha}^r$ and $\mu_{\alpha}$ are the relative permeability and viscosity of fluid phase $\alpha$, and ${p}_{\alpha}$ represents the pressure of fluid phase $\alpha$. 

The mass content of fluid phase $\alpha$ is expressed as 
\begin{align}
    m_{\alpha} = \rho_{\alpha} S_{\alpha} \phi(1+\varepsilon_v). \label{eq13}
\end{align}
Considering that the fluid content of a representative elementary volume evolves as a function of the change in volumetric strain, pore pressure and temperature, the variation of fluid content for a multiphase system is given as \cite{coussy2004poromechanics,jhajuanes14-wrr}
\begin{align}
{\left(\frac{\delta m}{\rho}\right)_{\alpha}} = {b_{\alpha}}{\varepsilon_v} + \sum_{k}{N_{jk}}{\delta p_{k}} - {\beta_{s,j}}{\delta T}, \label{eq14}
\end{align}
in which 
\begin{align}
 {N_{nn}} &= -{\phi}{\frac{\partial S_w}{\partial p_c}} + {\phi}{S_n}{c_n} + {S_n^2}{N}, \label{eq15} \\
 {N_{nw}} &= {N_{wn}} = {\phi}{\frac{\partial S_w}{\partial p_c}} + {S_n}{S_w}{N},  \label{eq16} \\
 {N_{ww}} &= -{\phi}{\frac{\partial S_w}{\partial p_c}} + {\phi}{S_w}{c_w} + {S_w^2}{N}, \label{eq17} \\
 N &= (b-\phi)(1-b)/{K_s}, \label{eq18} \\
 \beta_{s,j} &= S_{\alpha}((b-\phi)\beta_s+\phi\beta_{\alpha}), \label{eq19} 
\end{align} 
where ${c_{\alpha}}$ and ${\beta_{\alpha}}$ are compressibility and thermal expansion coefficients of fluid phase $\alpha$ as, respectively.

\subsection{Heat Transfer Constitutive Relations}
The energy content $m_\theta$ and the heat-flux $\boldsymbol{h}_{\theta}$ in \eqref{eq3} are expressed as 
\begin{align}
 {m_{\theta}} &= {(\rho C)_\text{avg}}{T},  \label{eq20} \\
 {\boldsymbol{h}_{\theta}} &= ({\rho_n}{C_n}{\boldsymbol{v}_n} + {\rho_w}{C_w}{\boldsymbol{v}_w}){T} -{\lambda_\text{avg}}{\boldsymbol{\nabla}}{T} \label{eq21},
\end{align}
where $(\rho C)_\text{avg}$~is the average heat capacity,
\begin{equation}
    (\rho C)_\text{avg}={(1-\phi)\rho_s C_s T} + \sum_{\alpha}{\phi S_{\alpha} \rho_{\alpha} C_{\alpha}}
\end{equation}
with ${C_s}$ the heat capacity of the solid phase, ${C_{\alpha}}$ the heat capacity of fluid phase $\alpha$, and ${\lambda_\text{avg}}$ is the average thermal conductivity of the porous medium,
\begin{align}
{\lambda_\text{avg}} = {(1-\phi)}{\lambda_s} + \sum_{\alpha}{\phi}{S_{\alpha}}{\lambda_{\alpha}}. \label{eq22}
\end{align}

\subsection{THM Governing Relations}
The linear momentum equation can be obtained by replacing \eqref{eq10} into \eqref{eq1} as 
\begin{align}
\begin{split}
& {K_{dr}}\boldsymbol{\nabla}{\varepsilon_v}+{\frac{1}{2}}{\nu^*}{K_{dr}} {\boldsymbol{\nabla} {(\boldsymbol{\nabla}{\cdot}{\boldsymbol{u}})}} + {\frac{3}{2}}{\nu^* K_{dr}}{\boldsymbol{\nabla}{\cdot}{(\boldsymbol{\nabla u})}}  \\
& \qquad\qquad\qquad\qquad - {b \sum_{\alpha}{\boldsymbol{\nabla}{(S_{\alpha} p_{\alpha})}}} - {\beta_s}{K_{dr}}{\boldsymbol{\nabla} {T}} + {\rho_b}{g}{\boldsymbol{d}} = {\boldsymbol{0}},
\end{split} \label{eq23}
\end{align}
where $\rho_b$~is the bulk density,
\begin{equation}
    \rho_b=(1-\phi)\rho_s+\phi(S_w\rho_w+S_n\rho_n).
\end{equation}
Substituting the constitutive relations \eqref{eq14} and \eqref{eq11} into \eqref{eq2}, yields the mass conservation equation for fluid phases in terms of volumetric stress as
\begin{align}
\begin{split}
&{\sum_{k}{(N_{jk}+\frac{b_{\alpha} b_k}{K_{dr}})}}{\frac{\partial p_k}{\partial t}} + {\frac{b S_{\alpha}}{K_{dr}}}{\frac{\partial \sigma_v}{\partial t}} - {({\beta_{s,j}} - {\beta_s b S_{\alpha}})}{\frac{\partial T}{\partial t}} \\
&~~~~~~~~~~- {\frac{k}{\mu_{\alpha}}}{\boldsymbol{\nabla}}{\cdot}{[{k_{\alpha}^r}({\boldsymbol{\nabla} p_{\alpha}} - {\rho_{\alpha}}{g}{\boldsymbol{d}})]} - {f_{\alpha}} = {0},
\end{split}
\label{eq24}
\end{align}
Finally, the governing equation for heat transfer is derived by replacing \cref{eq20,eq21} into \eqref{eq3}, as
\begin{align}
{(\rho C)_\text{avg}}\frac{\partial T}{\partial t} +{{(\rho_n C_n \boldsymbol{v}_n + \rho_w C_w \boldsymbol{v}_w) {\cdot}{\boldsymbol{\nabla}}{T}} - {\boldsymbol{\nabla}}{\cdot}({\lambda_\text{avg}}{\boldsymbol{\nabla}{T}}}) - {G_\theta} = {0}. \label{eq25}
\end{align}

\subsection{Dimensionless Governing Relations}
Considering dimensionless variables for wetting and nonwetting fluid flow under non-isothermal condition as
\begin{align}
{\bar{t}} = \frac{t}{t^*},~~~{\bar{x}} = \frac{x}{x^*},~~~{\boldsymbol{\bar{u}}} = \frac{\boldsymbol{u}}{u^*}, ~~~{\bar{\boldsymbol{\varepsilon}}} = \frac{\boldsymbol{\varepsilon}}{{\varepsilon}^*},~~~{\bar{p}}_{\alpha} = \frac{p_{\alpha}}{p^*},~~~{\bar{\boldsymbol{\sigma}}} = \frac{\boldsymbol{\sigma}}{p^*},~~~{\bar{T}} = \frac{T}{T^*}, \label{eq26}
\end{align}
The non-dimensional form of the linear momentum equation and stress-strain relation given in \eqref{eq23} and \eqref{eq10} are given by:
\begin{align}
& \boldsymbol{\bar{\nabla}}{\bar{\varepsilon}_v}+{\frac{1}{2}}{\nu^*}{\boldsymbol{\bar{\nabla}}}{(\boldsymbol{\bar{\nabla}}{\cdot}{\boldsymbol{\bar{u}}})} + \frac{3}{2}{\nu^*}{\boldsymbol{\bar{\nabla}}}{\cdot}{(\boldsymbol{\bar{\nabla}}{\boldsymbol{\bar{u}}})} - {b}{\sum_{\alpha}}{\boldsymbol{\bar{\nabla}}}{({S_{\alpha}}{\bar{p}_{\alpha}})} - {N_T}{\boldsymbol{\bar{\nabla}}}{\bar{T}} + {\frac{\rho_b}{\rho}}{N_d}{\boldsymbol{d}} = \boldsymbol{0}, \label{eq27} \\
& {\boldsymbol{\bar{\sigma}}-\boldsymbol{{\bar{\sigma}}_0}} = {\bar{\varepsilon}_v} {\mathbf{1}} + {3 \nu^*}{\boldsymbol{\bar{e}}} - {b}\sum_{\alpha}{S_{\alpha} ({\bar{p}_{\alpha}} - {{\bar{p}_{j0}}})} {\mathbf{1}} - {{N_T} {(\bar{T} - {\bar{T}}_0)}}{\mathbf{1}}, \label{eq28}  \\
& {\delta {\bar{\sigma}}_v} = {\delta {\bar{\varepsilon}}_v} -{b}{\sum_{\alpha}}{S_{\alpha}}{\delta{\bar{p}_{\alpha}}} - {N_T}{\delta {\bar{T}}}, \label{eq29}  
\end{align}
where we choose ${\rho}={(1-\phi)}{\rho_s}+{0.5}{\phi}({\rho_w}+{\rho_n})$. The dimensionless form of mass conservation for fluid phases \eqref{eq24} based on the volumetric stress can be expressed as:
\begin{align}
& {\sum_{k} ({N_{jk}}+\frac{b_{\alpha} b_k}{K_{dr}}){\bar{M}^*}{A^*}{\frac{\partial \bar{p}_k}{\partial \bar{t}}}} + {D_{\alpha}^*} \frac{\partial \bar{\sigma}_v}{\partial \bar{t}} - {Q_{\alpha}^*} \frac{\partial \bar{T}}{\partial \bar{t}} - {\frac{\mu}{\mu_{\alpha}}}{\boldsymbol{\bar{\nabla}}}{\cdot}{[{k_{\alpha}^r}{({\boldsymbol{\bar{\nabla}} {\bar{p}_{\alpha}}}-{\frac{\rho_{\alpha}}{\rho}}{N_d}{\boldsymbol{d}})}]} - {f_{\alpha}^*} = {{0}},  \label{eq30} 
\end{align}
in which ${\mu}$ is taken as ${\mu}={\mu_w}+{\mu_n}$. This relation can also be expressed using volumetric strain, however, it is not useful for us as we follow the sequential stress-split training. The dimensionless form of Darcy's law is expressed as:
\begin{align}
    \bar{\bs{v}}_{\alpha} = -\frac{\mu}{\mu_{\alpha}} k_{\alpha}^r \left({\boldsymbol{\bar{\nabla}} {\bar{p}_{\alpha}}}-{\frac{\rho_{\alpha}}{\rho}}{N_d}{\boldsymbol{d}}\right). \label{eq31}
\end{align}
The energy \cref{eq25} can be rewritten in the non-dimensional format as:
\begin{align}
{C^*}{\frac{\partial \bar{T}}{\partial \bar{t}}} + {J^*}[{\frac{\rho_n C_n}{\overline{\rho C}}}\bar{\bs{v}}_n
+ {\frac{\rho_w C_w}{\overline{\rho C}}}\bar{\bs{v}}_w] {\cdot}{\boldsymbol{\bar{\nabla}}}{\bar{T}}- {F^*}{\boldsymbol{\bar{\nabla}}}{\cdot}{({\frac{\lambda_\text{avg}}{\overline{\lambda}}}{\boldsymbol{\bar{\nabla}}}{\bar{T}})} - {G_\theta^*} = {0}, \label{eq32}
\end{align}
where the dimensionless parameters are
\begin{align}
\begin{split}
& {u^*} = {\frac{p^*}{K_{dr}}}{x^*},~ {\varepsilon^*} = {\frac{u^*}{x^*}}, ~{N_T} = {\beta_s}{K_{dr}}\frac{T^*}{p^*},~{N_d} = {\frac{x^* \rho}{p^*}}{g}, \\
& {\frac{1}{\bar{M}}} = {0.5}{\phi}{c_w}+{0.5}{\phi}{c_n}+\frac{b-\phi}{K_s},~ {\frac{1}{\bar{M}^*}} = {\frac{1}{\bar{M}}} +{\frac{b^2}{K_{dr}}}, ~ {A^*} = \frac{\mu {x^*}^2}{k {t^*}{\bar{M}^*}},\\
&{D_{\alpha}^*} = {S_{\alpha}}{D^*},~ {D^*}=\frac{b {\mu}{x^*}^2}{K_{dr} {k}{t^*}},~{Q_{\alpha}^*} = ({\beta_{s,j}} - {b S_{\alpha} \beta_s})\frac{T^* {\mu}{x^*}^2}{{t^*}{k}{p^*}}, ~{f_{\alpha}^*} = {f_{\alpha}}\frac{\mu {x^*}^2}{k p^*},\\
&{C^*} = {\frac{(\rho C)_\text{avg}}{\overline{\rho C}}},~ {J^*} = {\frac{k {p^*}{t^*}}{\mu {x^*}^2}}, ~ {F^*} = \frac{{t^*} \overline{\lambda}}{\overline{\rho C}{x^*}^2},~ {G_\theta^*} = {G_\theta}{\frac{{t^*}}{T^* \overline{\rho C}}}, \\
& {\overline{\rho C}} = {(1-\phi)}{\rho_s C_s} + {0.5}{\phi \rho_w C_w} + {0.5}{\phi \rho_n C_n}, ~ {\overline{\lambda}} = {(1-\phi)}{\lambda_s} + {0.5}{\phi \lambda_w}+{0.5}{\phi \lambda_n}.
\end{split}
\label{eq33}
\end{align}
This completes the description of the governing equations for THM coupled processes in a porous medium.

\section{PINN-Thermo-Hydro-Mechanical Framework}
Here, we first summarize the PINN framework for solving boundary value problems and then apply it to the THM problem governed by the equations in the previous section.

\subsection{Physics-Informed Neural Networks}
In the PINN framework, the unknown solution variables, e.g. temperature or displacement fields, are approximated using fully connected neural networks, with inputs as spatiotemporal variables (features) and final outputs as those field variables of interest \cite{IntroductionPINN}. For instance, the unknown temperature field $T(\bs{x},t)$ is approximated using a $L$-layer feed-forward neural network as 
\begin{align}
\hat{T}(\bs{x}, t) = {\Sigma^L}~{\circ}~{\Sigma^{L-1}}~{\circ}~{\cdots}~{\circ}~{\Sigma^1}{({\boldsymbol{x}},t)}, \label{eq34}
\end{align}
where $\hat{T}$ represents an approximation of $T$. $\Sigma^l$ represents a neural network layer, a nonlinear transformation of inputs, which is expressed mathematically as
\begin{align}
{\boldsymbol{\hat{y}}^l} = \Sigma^l( {\boldsymbol{\hat{x}}^{l-1}}) := {\sigma^l}{({\mathbf{W}^l}\cdot {\boldsymbol{\hat{x}}^{l-1}} + {\mathbf{b}^l})} \label{eq35}
\end{align}
where ${\boldsymbol{\hat{x}}^{l-1}}~{\in}~{\mathbb{R}^{I_l}}$ and ${\boldsymbol{\hat{y}}^l}~{\in}~{\mathbb{R}^{O_l}}$ are inputs to, and outputs from, layer $l$ with dimensions ${I_l}$ and ${O_l}$, respectively. $\boldsymbol{\hat{x}}^0$ constitutes the main inputs to the network, i.e., $(\bs{x}, t) \in \mathbb{R}^{D+1}$ in this case with $D$ as the spatial dimension, and $\boldsymbol{\hat{y}}^l$ is the final output from the network, i.e., $T \in \mathbb{R}$. 
${\sigma^l}~{:}~{\mathbb{R}}{\rightarrow}{\mathbb{R}}$ represents the activation function; it introduces nonlinearity into the network. Some common activation functions are Rectified Linear Unit (ReLU), Sigmoid, and Hyperbolic-Tangent (tanh) \cite{goodfellow2016deep}. ${\mathbf{W}^l}~{\in}~{\mathbb{R}^{{O_l}{\times}{I_l}}}$ and ${\mathbf{b}^l}~{\in}~{\mathbb{R}^{O_i}}$ are network parameters for layer ${l}$, also known as weights and biases, and are conveniently collected in  $\boldsymbol{\theta}^{l} \in \mathbb{R}^{N_l}$ with $N_l={O_l}{\times}{I_l} + O_i$ as the total number of parameters of layer $l$. For convenience, we express the fully-connected neural network, \cref{eq35}, as 
\begin{align}
    \hat{T}(\bs{x}, t) = \mc{N}_T(\bs{x}, t; \bs{\theta}), \label{eq36}
\end{align}
where $\bs{\theta} \in \mathbb{R}^N$ is the set of all $N$ parameters of the network. Since this construct forms a continuous approximation for the temperature field as a function of $\bs{x}, t$, i.e., $\hat{T}(\bs{x},t)$, we can leverage numerical or analytical differentiation to evaluate a PDE residual. While numerical differentiation is prone to error, particularly for deep networks, analytical differentiation is readily available using Automatic-Differentiation \cite{baydin2018automatic}, a built-in feature of modern DL frameworks such as \hbox{TensorFlow} \cite{abadi2016tensorflow}.

While data-driven training focuses on finding optimal values for network parameters such that network outputs match given data, Physics-Informed learning finds a set of parameters that not only match given data, but also satisfies the underlying physics (in the form of PDEs, initial and boundary conditions). Therefore, it is inherently a more challenging learning task, which results in a multi-objective optimization problem. Once trained, such a network tends to be more predictive than a purely data-driven one. 
For illustration, let us consider an initial and boundary value problem, given as 
\begin{equation}\label{eq37}
\begin{split}
{\mathcal{P}}(T)(\boldsymbol{x},t) - {f(\boldsymbol{x},t)} &= 0,~~~~~~~~~~~~~{\text{where}}~~~{\boldsymbol{x}}~{\in}~{\Omega},~~{t}~{\in}~{\Phi}, \\
{T(\tilde{\boldsymbol{x}},t)} &= {g(\tilde{\boldsymbol{x}},t)},~~~~~{\text{where}}~~~{\tilde{\boldsymbol{x}}}~{\in}~{\partial\Omega},~~{t}~{\in}~{\Phi}, \\ 
{T(\boldsymbol{x},t_0)} &= {h(\boldsymbol{x},t_0)},~~~{\text{where}}~~~{\boldsymbol{x}}~{\in}~{\Omega}, ~~t=t_0,
\end{split}
\end{equation}
in which ${\mathcal{P}}$ represents a partial differential operator, $T$ is the unknown temperature field,  ${f(\boldsymbol{x},t)}$ is a source term, and ${\Omega}~{\in}~{\mathbb{R}^D}$ and ${\Phi}~{\in}~{\mathbb{R}}$ denote spatial and temporal domains, respectively. 
The boundary conditions are specified by ${g(\tilde{\boldsymbol{x}},t)}$ with $\tilde{\boldsymbol{x}}$ as the boundary collocation points, and ${h(\boldsymbol{x},t_0)}$ represents the initial condition. 

In the PINN framework, the unknown variable $T$ is approximated using a neural network (\cref{eq36}).
The loss function is then constructed based on the linear combination of the above-mentioned constraints. In other words, it penalizes the network for deviation from the exact solution of PDEs with respect to boundary/initial conditions. Given ${\hat{T}(\boldsymbol{x},t)}$ as the approximate solution, the loss function is expressed as 
\begin{equation}\label{eq38}
\begin{split}
{\mathcal{L}}{(\boldsymbol{x},t;\boldsymbol{\theta})}~ & {=}~{\lambda_1}{\left \| {{\mathcal{P}}{\hat{T}(\boldsymbol{x},t)} - {f(\boldsymbol{x},t)}} \right \| }_{\Omega \times \Phi}\\
& {+}~{\lambda_2}{\left \| {{\hat{T}(\tilde{\boldsymbol{x}},t)} - {g(\tilde{\boldsymbol{x}},t)}} \right \|}_{\partial \Omega \times \Phi} \\
& {+}~{\lambda_3}{\left \| {{\hat{T}(\boldsymbol{x},t_0)} - {h(\boldsymbol{x},t_0)}} \right \|}_{\Omega \times \Phi_0}
\end{split}
\end{equation}
with ${\lambda_i}$ as the weight (penalty) of each term in the loss function. The notation $\left\|\circ\right\|$ represent the mean-squared error norm (MSE) commonly used for PINN solvers. The loss function represents the error in the approximation by the neural network. If the problem involves both Dirichlet and Neumann boundary conditions, the second term of the loss function consists of two separate terms with distinct weights for each one. 

As the last step, the solution to the desired boundary value problem \eqref{eq37} is identified by minimizing the total loss function \eqref{eq38} on a set of $M$ collocation points, $\mathbf{X} \in \mathbb{R}^{M\times D}, \mathbf{T}\in\mathbb{R}^{M\times 1}$. 
Consequently, the final optimization problem is expressed as:
\begin{align}
\boldsymbol{\theta}^* = \argmin_{\bs{\theta} \in \mathbb{R}^N}~{\mathcal{L}}{(\mathbf{X}, \mathbf{T}; \boldsymbol{\theta})}, \label{eq39}
\end{align}
where $\bs{\theta}^*$ represent the optimal network parameters. The collocation points $\mathbf{X}, \mathbf{T}$ are randomly selected inside the domain and on the boundary. Different optimization methods can be used \cite{nocedal2006numerical}, with the Adam optimizer being the most common \cite{kingma2014adam}. Both full-batch and mini-batch optimization strategies are applicable. In the latter, one must ensure that each mini-batch contains uniform samples for the full spatio-temporal domain. In addition, the loss penalty terms $\lambda_i$ can be selected based on adaptive strategies \cite{wang2022and,our-HM-paper}.

\subsection{PINN-THM Framework}
For the poroelasticity formulation above with two-phase immiscible flow under non-isothermal condition, the unknown solution variables in~2D are $u_x$, $u_y$, $p_w$, $p_n$, and $T$. As it is common in the FEM context, instead of $p_n$ we choose the capillary pressure $p_c=p_n-p_w$ as the unknown solution variable. Given that $p$ and $T$ are coupled with the volumetric strain $\varepsilon_v$ (\cref{eq29}), and also considering that the derivatives of multi-layer neural network takes very complicated forms, we also take the volumetric strain as an unknown so that we can better enforce the coupling between fluid pressure, temperature, and volumetric strains. This introduces an additional PDE for the volumetric strain, expressed as
\begin{align}
    \bar{\varepsilon}_v - \boldsymbol{\bar{\nabla}} \cdot \bar{\boldsymbol{u}} = 0. \label{eq40}
\end{align}
We find that this strategy, which has a long history in FEM modeling of quasi-incompressible materials, improves the training \cite{zienkiewicz1977finite,hughes2012finite}. Therefore, the neural networks for the dimensionless form of these variables are,
\begin{align}
    \bar{u}_{\bar{x}} &: (\bar{x}, \bar{y}, \bar{t}) \mapsto \mathcal{N}_{\bar{u}_{\bar{x}}}(\bar{x}, \bar{y}, \bar{t}; \boldsymbol{\theta}_{\bar{u}_{\bar{x}}}), \label{eq41} \\
    \bar{u}_{\bar{y}} &: (\bar{x}, \bar{y}, \bar{t}) \mapsto \mathcal{N}_{\bar{u}_{\bar{y}}}(\bar{x}, \bar{y}, \bar{t}; \boldsymbol{\theta}_{\bar{u}_{\bar{y}}}), \label{eq42} \\
    \bar{p}_w &: (\bar{x}, \bar{y}, \bar{t}) \mapsto \mathcal{N}_{\bar{p}_w}(\bar{x}, \bar{y}, \bar{t}; \boldsymbol{\theta}_{\bar{p}_w}), \label{eq43} \\
    \bar{p}_c &: (\bar{x}, \bar{y}, \bar{t}) \mapsto \mathcal{N}_{\bar{p}_c}(\bar{x}, \bar{y}, \bar{t}; \boldsymbol{\theta}_{\bar{p}_c}), \label{eq44} \\
    \bar{\varepsilon}_v &: (\bar{x}, \bar{y}, \bar{t}) \mapsto \mathcal{N}_{\bar{\varepsilon}_v}(\bar{x}, \bar{y}, \bar{t}; \boldsymbol{\theta}_{\bar{\varepsilon}_v}), \label{eq45} \\
     \bar{T} &: (\bar{x}, \bar{y}, \bar{t}) \mapsto \mathcal{N}_{\bar{T}}(\bar{x}, \bar{y}, \bar{t}; \boldsymbol{\theta}_{\bar{T}}), \label{eq46} 
\end{align}
where $\boldsymbol{\theta}_\alpha$ highlights that these networks have independent parameters. Given the general PDEs of THM modeling in \cref{eq27,eq30,eq32}, below we summarize the loss functions for 1D~problems, to avoid very long equations. Considering the three coupled constituents, i.e., matrix deformation, two phase flow relations, and heat transfer, we end up having three loss functions associated with each of these, as 
\begin{equation}
\begin{split}
\mathcal{L}_s &= \lambda_1 \left\| \frac{\partial \bar{\varepsilon}_{v}}{\partial \bar{x}} + 2\nu^*\left(\frac{\partial^2 \bar{u}_{\bar{x}}}{\partial\bar{x}^2} \right) - b\frac{\partial (S_n\bar{p}_n + S_w\bar{p}_w)}{\partial \bar{x}} - {N_T}\frac{\partial \bar{T}}{\partial \bar{x}} + \frac{\rho_b}{\rho}N_d~d_{\bar{x}} \right\| \\
&+ \lambda_2 \left\| \bar{u}_{\bar{x}} - \tilde{u}_{\bar{x}} \right\| + \lambda_3 \left\| \bar{\sigma}_{\bar{xx}} - \tilde{\sigma}_{\bar{xx}} \right\| ,  \\
\mathcal{L}_f &= \lambda_4 \left\|\sum_k (N_{wk} + \frac{b_w b_k}{K_{dr}})\bar{M}^*{A}^*\frac{\partial\bar{p}_k}{\partial\bar{t}} + D^*_w \frac{\partial\bar{\sigma}_v}{\partial \bar{t}} - {Q}^*_w \frac{\partial\bar{T}}{\partial \bar{t}} - \frac{\mu}{\mu_w}k^r_w \left( \frac{\partial^2 \bar{p}_w}{\partial \bar{x}^2} \right) - f^*_w\right\| \\
&+ \lambda_{5} \left\|\sum_k (N_{nk} + \frac{b_n b_k}{K_{dr}})\bar{M}^*{A}^*\frac{\partial\bar{p}_k}{\partial\bar{t}} + D^*_n \frac{\partial\bar{\sigma}_v}{\partial \bar{t}} - {Q}^*_n \frac{\partial\bar{T}}{\partial \bar{t}} - \frac{\mu}{\mu_n}k^r_n \left( \frac{\partial^2 \bar{p}_n}{\partial \bar{x}^2} \right) - f^*_n\right\| \\
&+ \lambda_{6} \left\| \bar{p}_w - \tilde{p}_w \right\| 
+ \lambda_{7} \left\| \bar{p}_n - \tilde{p}_n \right\| 
+ \lambda_{8} \left\| \bar{q}_{\bar{x}}^{w} - \tilde{q}_{\bar{x}}^{w} \right\| 
+ \lambda_{9} \left\| \bar{q}_{\bar{x}}^{n} - \tilde{q}_{\bar{x}}^{n} \right\|, \\
\mathcal{L}_T &= \lambda_{10} \left\|C^*\frac{\partial\bar{T}}{\partial\bar{t}} + {J^*} 
\left[
\frac{\rho_n C_n}{\overline{\rho C}} \bar{v}^n_x+\frac{\rho_w C_w}{\overline{\rho C}} \bar{v}^w_x
\right]
\frac{\partial \bar{T}}{\partial \bar{x}} 
- {F}^* \frac{\partial}{\partial \bar{x}}{\left( \frac{\lambda_\text{avg}}{\bar{\lambda}} \frac{\partial \bar{T}}{\partial \bar{x}}  \right)} - {G_\theta^*} \right\| \\
& + \lambda_{11} \left\| \bar{T} - \hat{T} \right \| 
+ \lambda_{12} \left\| \bar{q}_{\bar{x}}^{T} - \tilde{q}_{\bar{x}}^{T} \right\|,
\end{split}
\label{eq47}
\end{equation}
and the total loss is evaluated as the summation of these three terms,
\begin{align}
\mc{L} = \mc{L}_s + \mc{L}_f + \mc{L}_T. \label{eq48}
\end{align}
Here, $\tilde{\circ}$ refers to the boundary or initial values for each term. Optimizing the total loss function, \cref{eq48}, is referred as the \emph{simultaneous solution} strategy. We find, however, that this is a challenging many-objective optimization problem with many hyper-parameters to tune, which did not lead to successful results for the problems reported in the next section. 

An alternative solution strategy is to train each loss term individually and iterate sequentially until convergence. This is a common solution strategy in the FEM community, where different operator splits can be adopted depending on which problem is solved first \cite{armero1992new,kimtchelepijuanes11-drained,kim-fixedStress,kim2018unconditionally}. Here, we follow our earlier work on the use of sequential-stress-split. Therefore, in the most general case, we first solve for temperature, followed by solving the two-phase flow equations to evaluate the pressure of the wetting and nonwetting phases, and finally solving the linear momentum equation for the displacements. The algorithm is summarized in \Cref{alg1}. While this strategy worked well for the problems explored in the next section, there could be problems that would benefit from alternative training and optimization strategies. 

\begin{algorithm}
\caption{Sequential fixed-stress-split algorithm for PINN-THM framework}\label{alg:cap}
\begin{algorithmic}[1]
\State $\mathbf{X}, \mathbf{T} \gets $ Sample uniformly spatial and temporal domains. 
\State $\bs{\theta}^n_{\bar{T}},\bs{\theta}^n_{\bar{p}_w},\bs{\theta}^n_{\bar{p}_c}, \bs{\theta}^n_{\bar{u}_{\bar{x}}},\bs{\theta}^n_{\bar{u}_{\bar{y}}},\bs{\theta}^n_{\bar{\varepsilon_v}} \gets $ Initialize randomly using Glorot scheme. 
\State $n \gets 1$
\State $\bar{\sigma}_v^0, \bar{T}^0 \gets \bs{0}$ 
\While{$ \text{err} > \text{TOL}$}
    \State $\bs{\theta}^n_{\bar{T}} \gets$ Optimize $\mc{L}_T$ for $\bs{\theta}_{\bar{T}}$ over $\mathbf{X}, \mathbf{T}, \bar{p}^{n-1}$.
    \State $\bar{T}^n \gets $ Evaluate $\bar{T}$ using $\bs{\theta}^n_{\bar{T}}$. 
    \State $\bs{\theta}^n_{\bar{p}_w}, \bs{\theta}^n_{\bar{p}_c} \gets$ Optimize $\mc{L}_f$ for $\bs{\theta}_{\bar{p}_w}, \bs{\theta}_{\bar{p}_c}$ over $\mathbf{X}, \mathbf{T}, \bar{\sigma}_v^{n-1}, \bar{T}^{n}$.
    \State $\bar{p}_w^n, \bar{p}_c^n \gets $ Evaluate $\bar{p}_w, \bar{p}_c$ using $\bs{\theta}^n_{\bar{p}_w}, \bs{\theta}^n_{\bar{p}_c}$. 
    \State $\bs{\theta}^n_{\bar{u}_{\bar{x}}},\bs{\theta}^n_{\bar{u}_{\bar{y}}},\bs{\theta}^n_{\bar{\varepsilon}_v} \gets $ Optimize $\mc{L}_s$ for $\bs{\theta}_{\bar{u}_{\bar{x}}},\bs{\theta}_{\bar{u}_{\bar{y}}},\bs{\theta}_{\bar{\varepsilon}_v}$ over $\mathbf{X}, \mathbf{T}, \bar{p}^{n}, \bar{T}^{n}$. 
    \State $\bar{\sigma}_v^n \gets $ Evaluate $\bar{\sigma}_v$ using $\bs{\theta}^n_{\bar{u}_{\bar{x}}},\bs{\theta}^n_{\bar{u}_{\bar{y}}},\bs{\theta}^n_{\bar{\varepsilon}_v}$. 
    \State $\bs{\theta}^n \gets \{\bs{\theta}^n_{\bar{T}}, \bs{\theta}^n_{\bar{p}_w}, \bs{\theta}^n_{\bar{p}_c},   \bs{\theta}^n_{\bar{u}_{\bar{x}}},\bs{\theta}^n_{\bar{u}_{\bar{y}}},\bs{\theta}^n_{\bar{\varepsilon}_v} \}$ as the collection of all parameters. 
    \State $\text{err} \gets \|\bs{\theta}^n - \bs{\theta}^{n-1}\| / \|\bs{\theta}^n\|$ with $\| \circ \|$ as the $L^2$ norm. 
    \State $n \gets n+1$
\EndWhile
\end{algorithmic}
\label{alg1}
\end{algorithm}

\section{Applications}
In a recent study, we validated the sequential PINN-poroelasticity solver for modeling isothermal fluid flow in deformable porous media by simulating various reference problems \cite{our-HM-paper}. Here, we focus on the additional energy equation, and consider four reference problems:
\begin{enumerate}
    \item One-dimensional heat-conduction in a soil-column in which we solve for temperature only.
    \item One-dimensional non-isothermal consolidation under fully saturated conditions by solving for temperature, pressure, and displacement. 
    \item One-dimensional non-isothermal consolidation under unsaturated conditions, by solving the full set of coupled equations, as summarized in \Cref{alg1}.
    \item Two-dimensional non-isothermal injection-production in a rigid fully-saturated  domain, by solving for temperature and pressure.
\end{enumerate}

The problems are all solved using SciANN \cite{haghighat2021sciann}, a Keras/TensorFlow API for physics-informed machine learning, developed by the authors, and shared in SciANN's github repository. \footnote{\href{https://github.com/sciann/sciann-applications}{https://github.com/sciann/sciann-applications}}


\subsection{Conduction With and Without Convection}
\begin{figure}[t]
    \centering
    \includegraphics[width=0.18\textwidth]{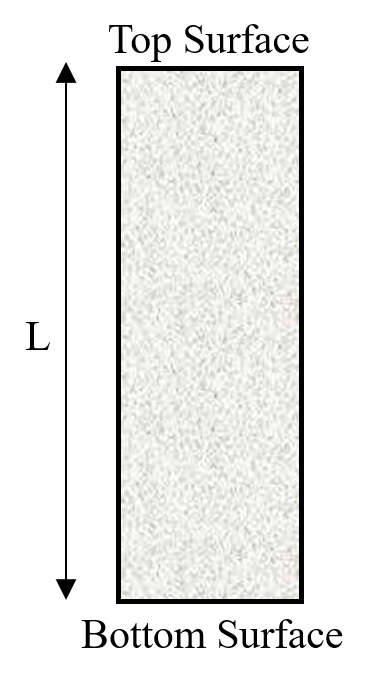}\caption{One-dimensional soil column with impervious, isolated, and displacement-free side faces.}
    \label{fig:example1-geometry}
\end{figure}

 As the first problem, let us consider a ${0.1~ \text{m}}$ saturated column of soil, which has been previously considered by \citet{20-THM-pak}, as shown in \Cref{fig:example1-geometry}. For simplification, we ignore the solid deformation ($E=\infty$), also we assume incompressible solid and fluid phases, i.e., ${{K_s}={K_w}=\infty}$, and neglect the gravity term. As material properties, the porosity and Biot coefficient are ${\phi=0.5}$, ${b=1}$. The intrinsic permeability is set to ${k={1.98}{\times}{10^{-5}~{\text{m}^2}}}$. The fluid viscosity is taken as $\mu_f = {{10^{-3}}~{\text{Pa} \cdot \text{s}}}$. The density of solid and fluid phases are ${{\rho_s}={2000}~{\text{kg}/\text{m}^3}}$, ${{\rho_f}={1000}~{\text{kg}/\text{m}^3}}$. The homogeneous thermal conductivity of the medium is taken as ${{\lambda_\text{avg}}={2.6}~{\text{J}/\text{s} \cdot \text{m} \cdot {^\circ}\text{C}}}$, and the heat capacity as ${{C_s}={C_f}={0.201}~{\text{J}/\text{kg} \cdot {^\circ}\text{C} }}$. The effect of thermal expansion coefficient is ignored, i.e., ${{\beta_s}={\beta_f}={0}}$. 

To assess the impact of convection and conduction on heat transfer, we consider three cases. Initial temperature and pressure for all cases are set to ${{T=0}~{^\circ}\text{C}}$, ${{p=0}~\text{Pa}}$. For all cases, the boundary conditions are ${{T=100}~{^\circ}\text{C}}$ at the top surface and ${{T=0}~{^\circ}\text{C}}$ at the bottom surface. For case (I), the pressure boundary conditions at both ends of the column are set as ${{p=0}~\text{Pa}}$, which implies that heat transfer occurs via conduction only (no fluid flow). For case (II), the top surface is set to ${{p=1}~\text{Pa}}$ and the bottom face is fixed at ${{p=0}~\text{Pa}}$, in which, convection helps the conduction to transfer more energy. For case (III), the top pressure is set to ${{p=0}~\text{Pa}}$ while the bottom pressure is ${{p=1}~\text{Pa}}$, in which, convection and conduction compete to neutralize each other.

In this example, due to the relatively high permeability of the medium, the fluid flow reaches the steady-state conditions quickly, therefore it can be considered time-independent,
and pressure takes a linear distribution. Therefore, here, we only solve the energy equation with a constant pressure gradient and the results are compared with those obtained by a classical FEM solution using COMSOL.

\begin{figure}[H]
    \centering
    \includegraphics[width=0.5\textwidth]{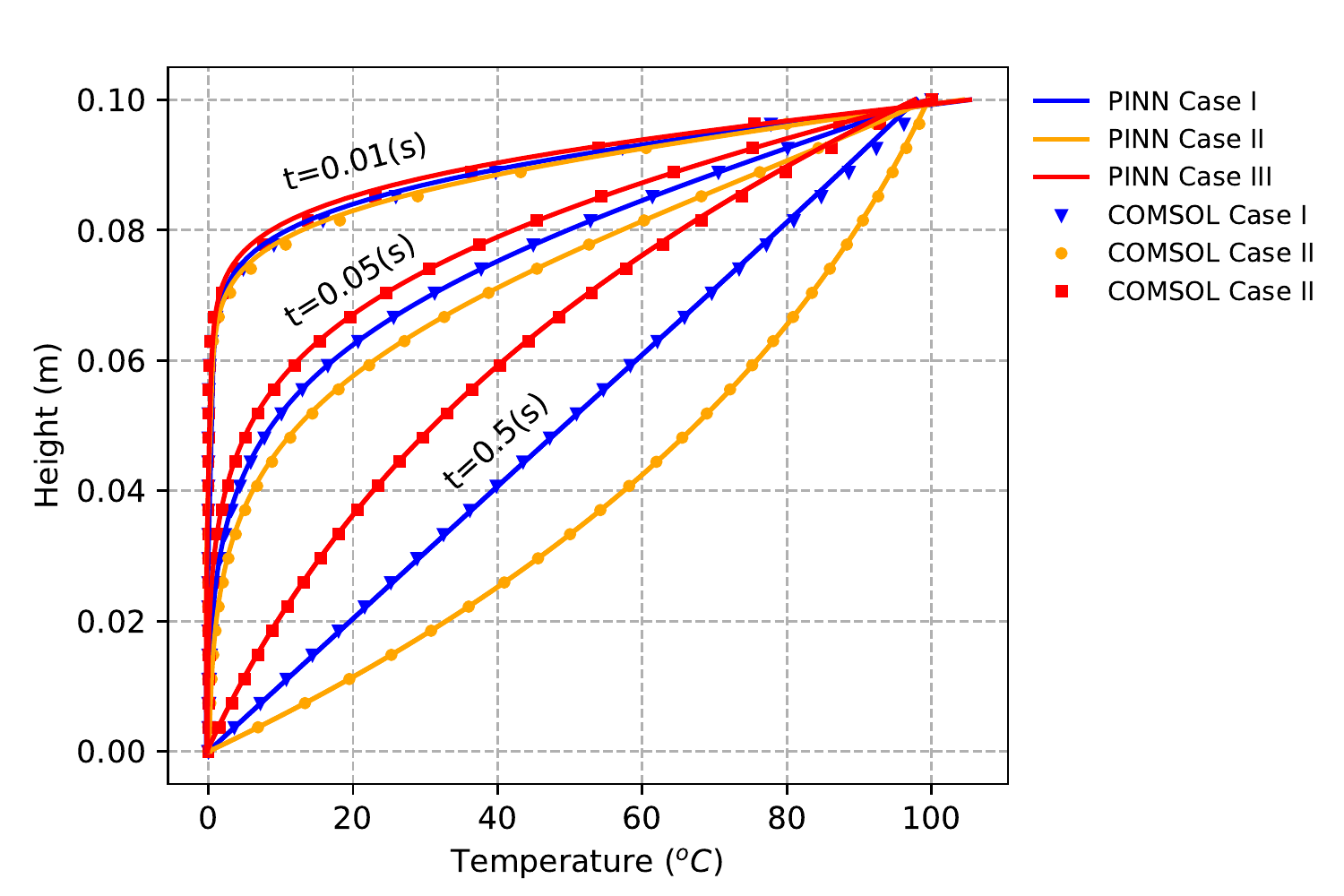}\caption{PINN's temperature solution (solid lines) vs. COMSOL's solution (symbols) as a function of height at different times, for the conduction problem with and without convection. }
    \label{fig:example1-LinePlot}
\end{figure}

\Cref{fig:example1-LinePlot} shows the distribution of temperature over the sample's height for all three cases at different times. In case (I), conduction creates a constant heat gradient in the domain. In case (II), convection accelerates conduction to facilitate energy transfer, so the average temperature is higher than in case (I). In case (III), convection opposes the energy transfer through conduction, so the temperature is lower than in the other cases. The results confirm that the energy relation is correctly implemented and PINN results are accurate. 

\subsection{Non-isothermal Consolidation of a Saturated Soil Column}
As a second problem, we model the consolidation of a saturated soil column under non-isothermal conditions. This is one of the reference problems that has been considered extensively for validating numerical solvers \cite{3-THM-1984, Lewis-book, 15-THM-2018}. Here, the reference solution is generated based on the results by \citet{Lewis-book}. 
Solid and fluid phases are considered incompressible, and the gravity term is ignored. The soil column (see \Cref{fig:example1-geometry}) has a height of ${{7}~\text{m}}$, subjected to the initial conditions ${{T=0}~{^\circ}\text{C}}$, ${{p=0}~\text{Pa}}$, ${{u=0}~\text{m}}$.  
The boundary conditions include the sudden application of a compressive stress  ${{1000}~\text{Pa}}$ on the top surface. Furthermore, the temperature and pressure at the top surface are prescribed as ${{T=50}~{^\circ}\text{C}}$ and ${{p=0}~\text{Pa}}$. Heat flux, fluid flow, and the displacement of the bottom surface of the column are prescribed to be zero. These boundary conditions simulate the drainage of the sample from its top surface after loading. 

The material properties of the solid phase include  $E=6\times10^{6}~\text{Pa}$, ${{\nu}={0.4}}$, ${\phi=0.5}$, and ${b=1}$. The fluid flow parameters are given as ${k={4.64}{\times}{10^{-17}~{\text{m}^2}}}$ and ${{\mu_f}={10^{-3}}~{\text{Pa} \cdot \text{s}}}$. The solid and fluid densities are ${{\rho_s}={2000}~{\text{kg}/\text{m}^3}}$ and ${{\rho_f}={1000}~{\text{kg}/\text{m}^3}}$, respectively. The heat transfer parameters are considered as ${{\lambda_\text{avg}}={9.68}{\times}{10^{-3}}~{\text{J}/\text{s} \cdot \text{m} \cdot {^\circ}\text{C}}}$, ${({\rho C}_\text{avg})}={167360}~{\text{J}/ {\text{m}^3} \cdot {^\circ}\text{C}}$, ${{\beta_s}={9}{\times}{10^{-7}}~{1/{^\circ}\text{C}}}$, ${{\beta_s}={0.0}~{1/{^\circ}\text{C}}}$, as detailed by  \citet{Lewis-book}.

\begin{figure}[H]
    \centering
    \includegraphics[width=0.45\textwidth]{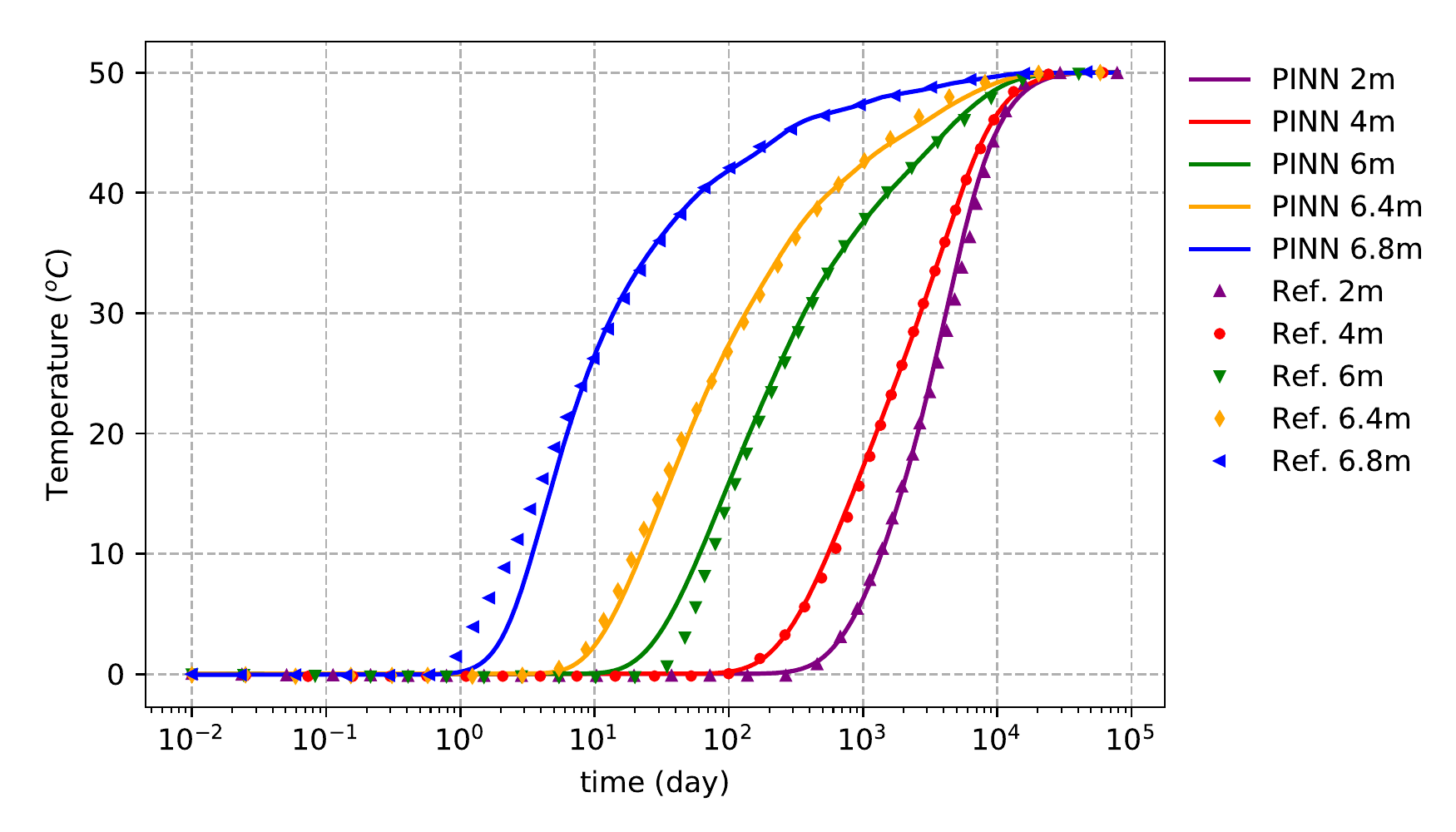}
    \includegraphics[width=0.45\textwidth]{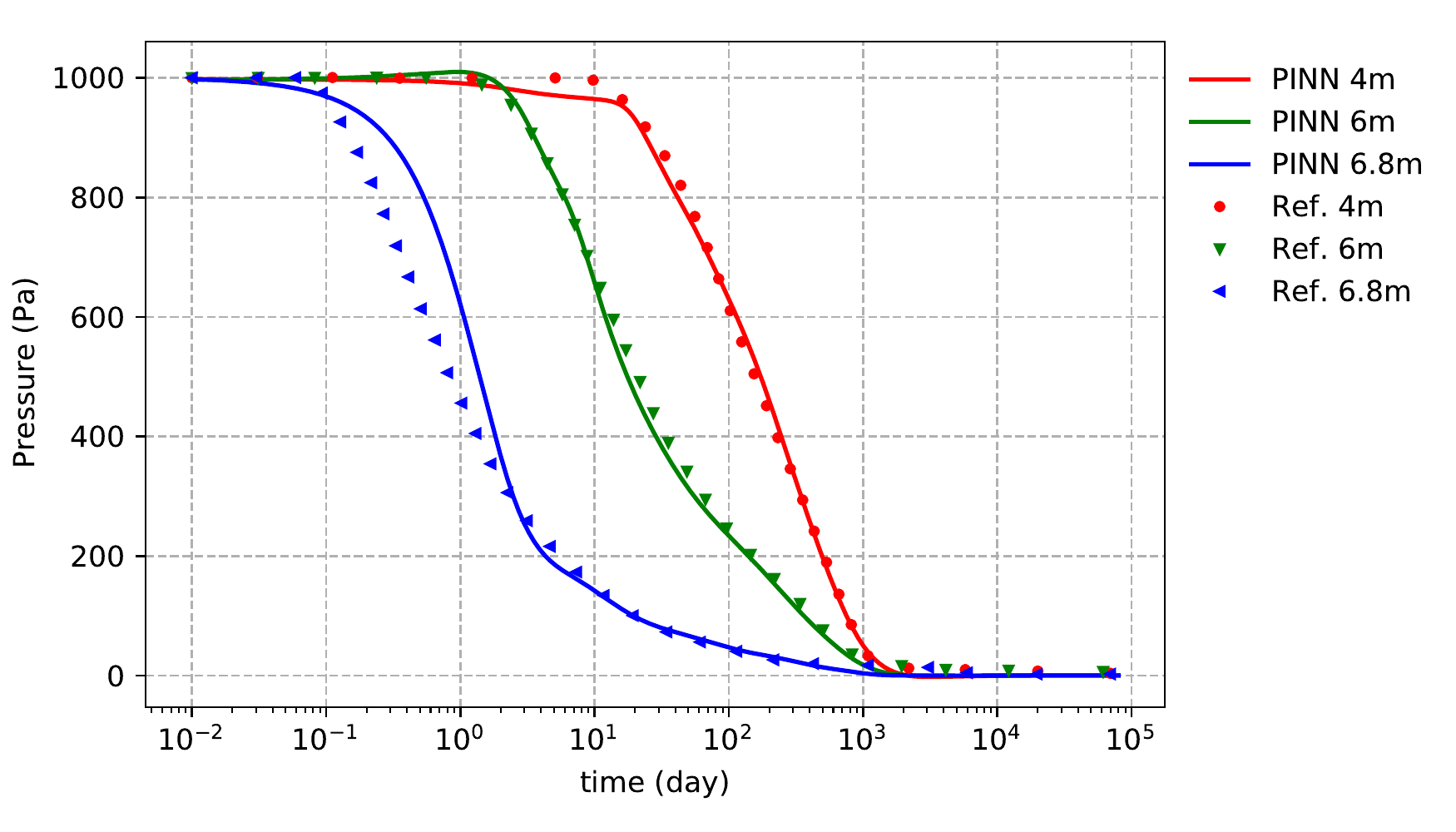}
    \includegraphics[width=0.45\textwidth]{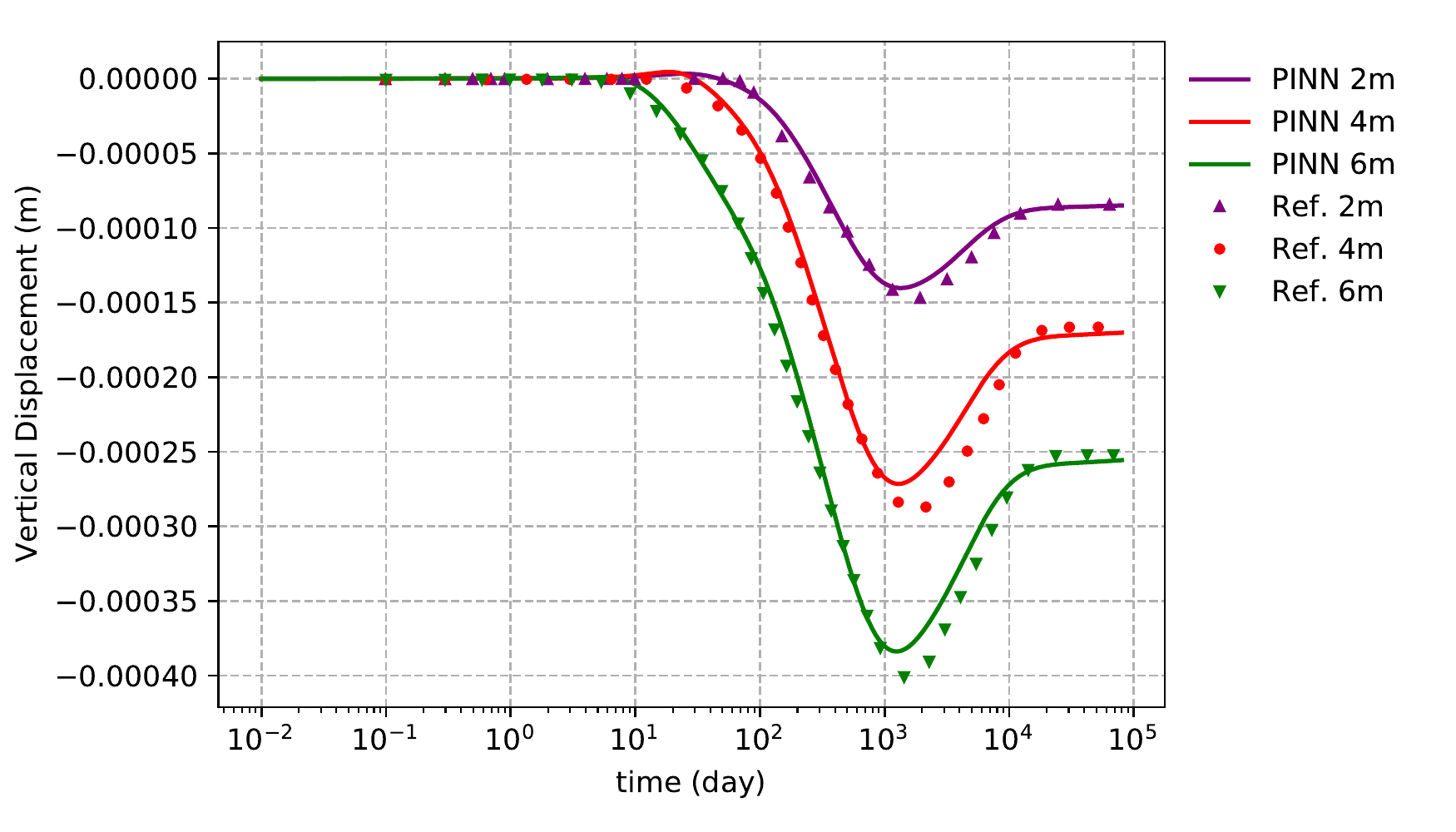}
    \caption{PINN solutions, i.e., temperature, pressure, and vertical displacement plots (solid lines), and comparison against the reference solution reported by \cite{Lewis-book} (symbols) for the problem of non-isothermal consolidation through a saturated soil column. The horizontal axis indicates time, and each color represents a different depth. }
    \label{fig:example2-temperature}
\end{figure}

The PINN solutions are plotted in \Cref{fig:example2-temperature}, with the heights calculated from the bottom of the column, which are in agreement with those reported by \citet{Lewis-book}. The temperature and pressure are both subject to a sudden change at the initial time. The pressure rises quickly across the sample, while the temperature increases more gradually. As the increased pressure dissipates from the sample and before the temperature rises to its maximum value, there is a time at which compaction is maximum (around 1,000~days), followed by a thermal rebound over time until steady-state conditions are reached. Therefore, before 1,000 days, the deformation process is dominated by the applied stress and pore pressure dissipation, while afterwards it is mostly controlled by the temperature increase. 

\subsection{Thermoelastic Consolidation of an Unsaturated Stratum}
As the next problem, let us consider thermoelastic consolidation of an unsaturated subsurface stratum, which includes two-phase fluid flow under non-isothermal conditions in deformable porous media \cite{8-THM-1996, Lewis-book, 14-THM-1981, 13-THM}. \citet{14-THM-1981} investigated the influence of various parameters including fluid compressibility moduli, air phase constitutive relation, thermal parameters, and permeability. They also modeled it under both isothermal and non-isothermal conditions and considering consolidation and swelling. \citet{8-THM-1996} considered phase change, and they modeled the consolidation of the stratum for different types of boundary conditions. \citet{13-THM} simulated consolidation and swelling conditions, without phase-change considerations. Here, we use the setup reported by \cite{khoei2021modeling} without including the phase-change effects.

We consider the top  ${{10}~\text{cm}}$ of a stratum, i.e., $L={{10}~\text{cm}}$, which experiences environmental changes that include a temperature increase of  ${{15}~{^\circ}\text{C}}$ and a capillary pressure increase of ${{140}~\text{kPa}}$. These sudden changes result in the heat and mass transfer in the porous medium similar to a drying process. The initial conditions for the  stratum capillary pressure, water pressure, and temperature include ${{p_c=280}~\text{kPa}}$, ${{p_g=102}~\text{kPa}}$, ${{T=10}~{^\circ}\text{C}}$, respectively, which are in equilibrium with the solid phase stresses. The boundary conditions include ${{p_c=420}~\text{kPa}}$, ${{p_g=102}~\text{kPa}}$, and ${{T=25}~{^\circ}\text{C}}$ at the top surface, due to the environmental changes. The bottom surface is considered impermeable, and its displacement is fixed. Gravity effects are also ignored here. 

Material properties include the elastic modulus ${E=60~\text{MPa}}$, the Poisson ratio ${{\nu}={0.2857}}$, the porosity ${\phi=0.5}$, and Biot modulus ${b=1}$. The bulk moduli are considered as ${K_s}=0.14 \times 10^{10}~\text{Pa}$, ${K_w}=0.43 \times 10^{13}~\text{Pa}$, and $K_g=0.1 \times 10^{6}~\text{Pa}$ for solid, water, and gas phases, respectively. The fluid flow parameters are taken as $k={6}{\times}{10^{-15}~{\text{m}^2}}$ and ${\mu_w}={\mu_g}={10^{-3}}~{\text{Pa}\cdot \text{s}}$. The densities  are ${\rho_s}={1800}~{\text{kg}/\text{m}^3}$, ${\rho_w}={1000}~{\text{kg}/\text{m}^3}$, and ${\rho_g}={1.22}~{\text{kg}/\text{m}^3}$. The thermal conductivity coefficient is considered as ${\lambda_\text{avg}}={0.458}~{\text{J}/\text{s} \cdot \text{m} \cdot {^\circ}\text{C}}$, with the heat capacity as ${C_s}={125460}~{\text{J}/\text{kg} \cdot {^\circ}\text{C}}$, ${C_w}={4182}~{\text{J}/\text{kg} \cdot {^\circ}\text{C}}$, and ${C_g}={1000}~{\text{J}/\text{kg} \cdot {^\circ}\text{C}}$. In addition, the thermal expansion coefficients are given as ${\beta_s}={9}{\times}{10^{-7}}~{1/{^\circ}\text{C}}$, ${\beta_w}={6.3}{\times}{10^{-6}}~{1/{^\circ}\text{C}}$, and ${\beta_g}={3.3}{\times}{10^{-3}}~{1/{^\circ}\text{C}}$, as considered by \citet{khoei2021modeling}. 
To describe the flow of two immiscible fluids through the porous media, the Brooks-Corey \cite{Brooks-Corey} relations for saturation and relative permeability of fluid phases are used:
\begin{align*}
    {p_c}={p_b}{S_e}^{-1/\lambda}, \quad {k_{rw}}={S_e}^{(2+3\lambda)/\lambda}, \quad 
    {k_{rg}}={(1-{S_e})^2}{(1-{S_e}^{(2+\lambda)/\lambda})}, \quad  {S_e}={\frac{S_w - S_{rw}}{1-{S_{rw}}}}, 
\end{align*}
in which ${S_e}$ denotes the effective saturation, and ${k_{rw}}$ and ${k_{rg}}$ are the relative permeabilities of water and gas phases, respectively. Finally, the pore size distribution and the residual water saturation parameters are ${\lambda=2.308}$ and ${S_{rw}=0.3216}$, respectively, and the capillary entry pressure is taken as ${{p_b=133.813}~\text{kPa}}$.

\begin{figure}[H]
    \centering
    \includegraphics[width=0.8\textwidth]{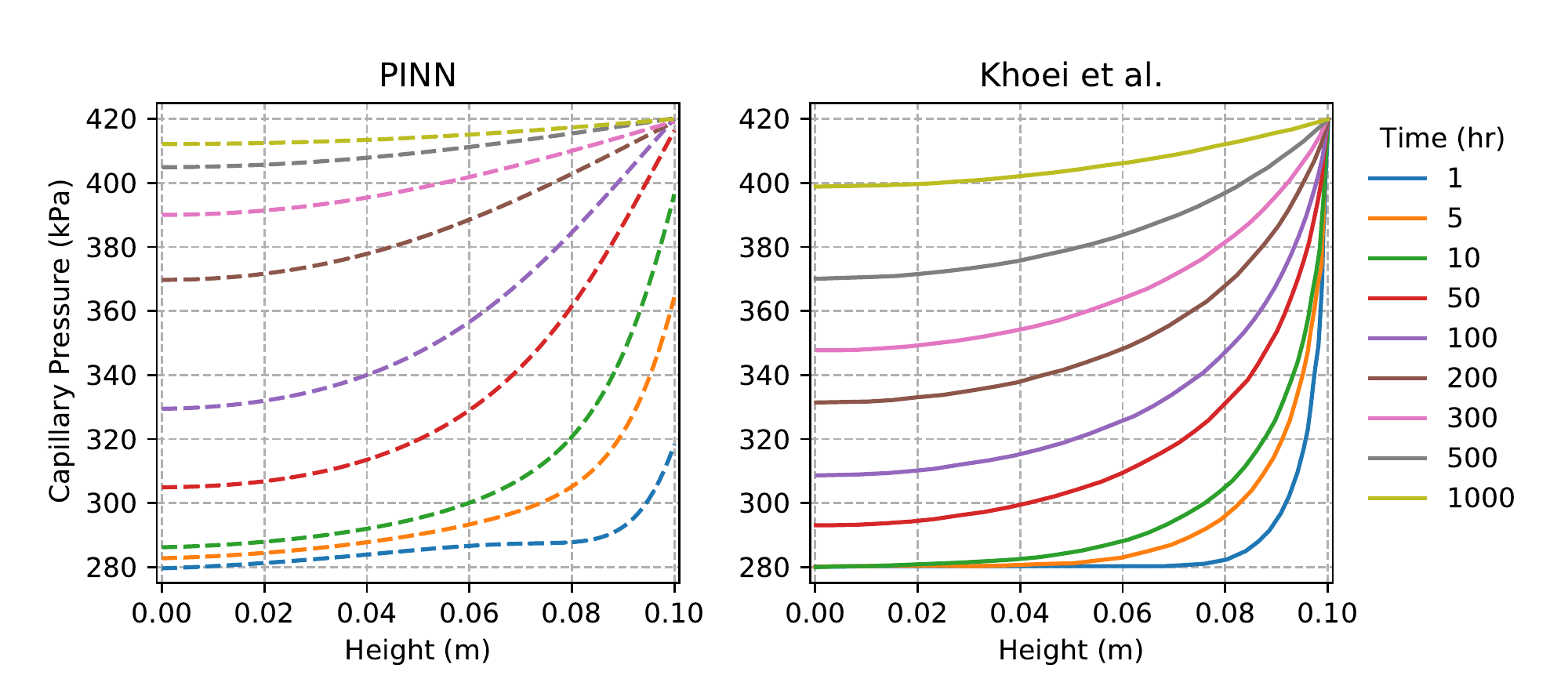}
    \includegraphics[width=0.8\textwidth]{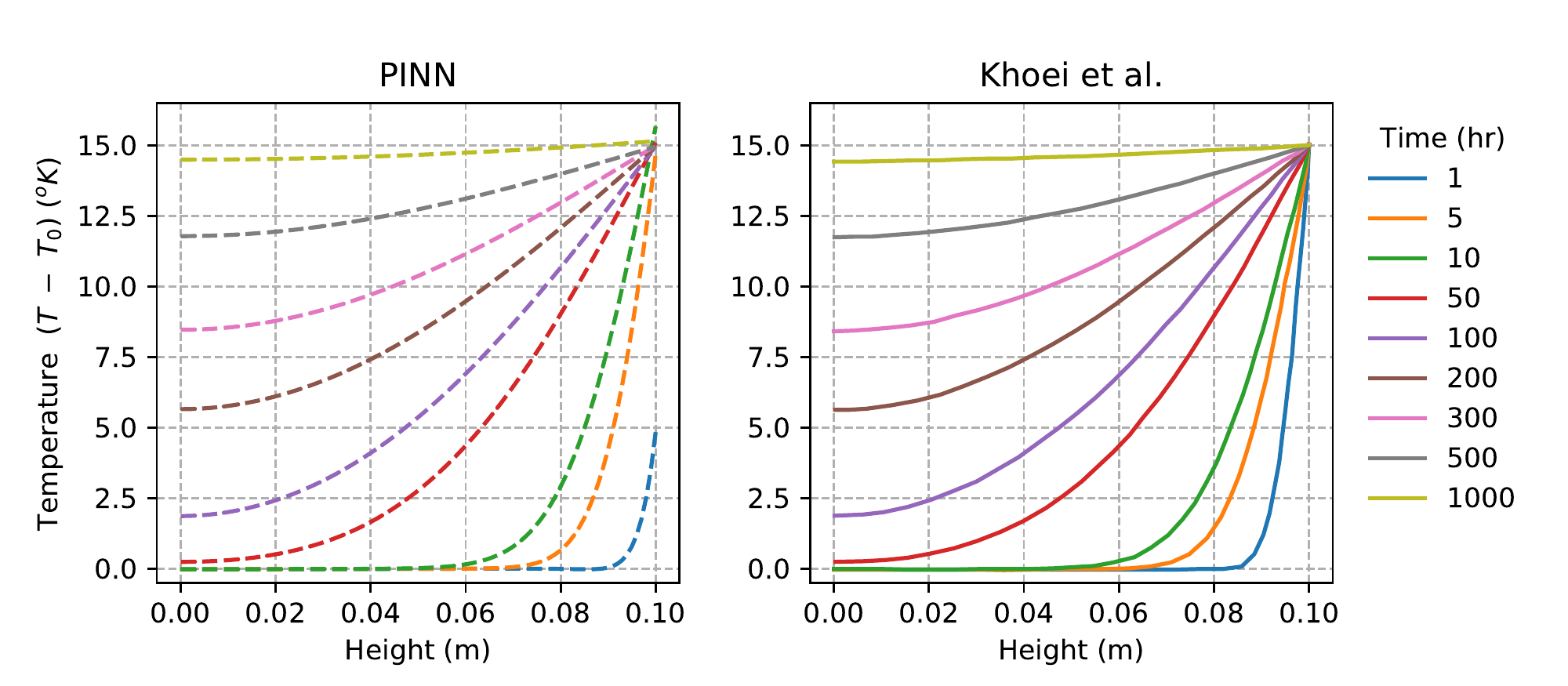}
    \includegraphics[width=0.8\textwidth]{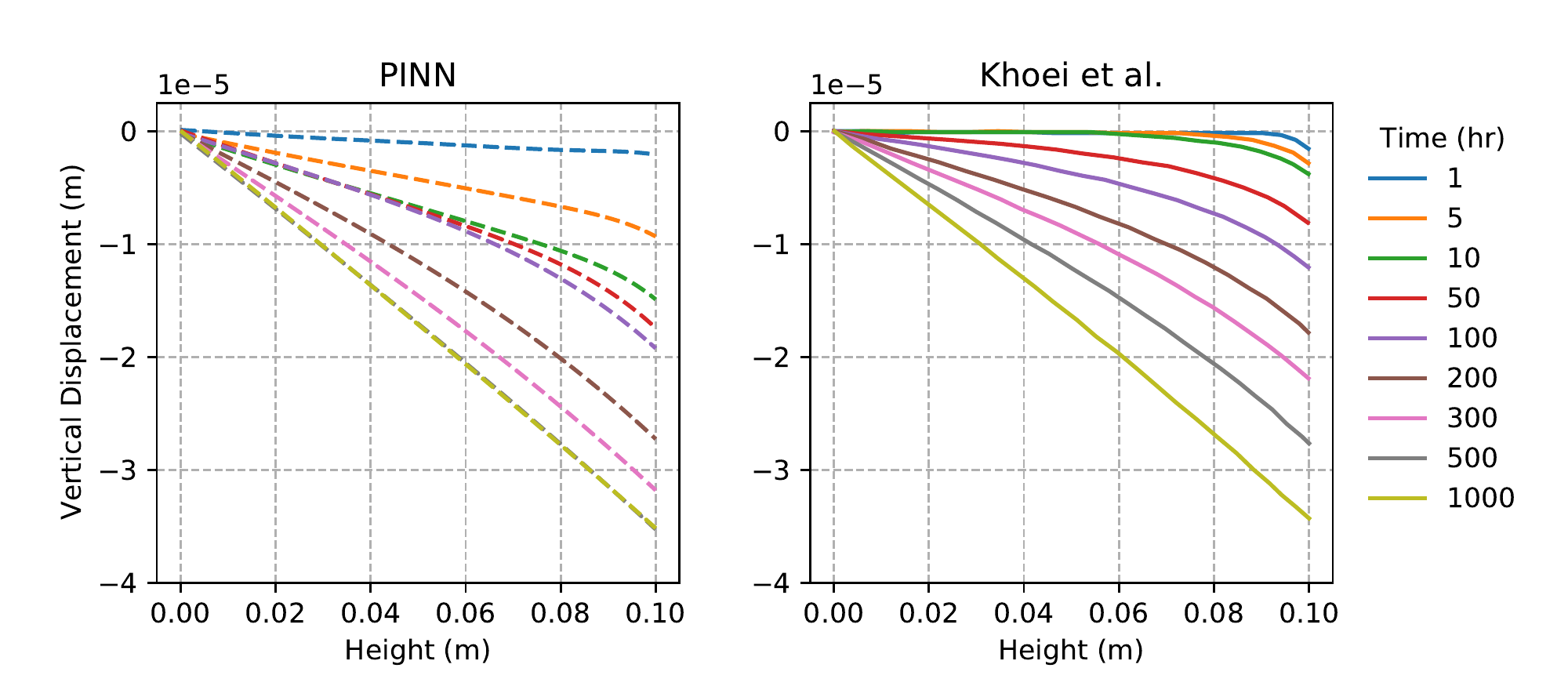}
    \caption{PINN solutions, i.e., capillary pressure, temperature, and vertical displacement (left column, dashed lines) and comparison against the FEM solutions reported by \citet{khoei2021modeling} (right column, solid lines) for the problem of thermoelastic consolidation of an unsaturated stratum. The horizontal axis indicates the depth of the stratum. Different colors show the solution at different times.}
    \label{fig:example3-Capillary}
\end{figure}

The temperature, capillary pressure, saturation, and displacement profiles at different times are shown in \Cref{fig:example3-Capillary}. Because the evolutions of heat transfer and fluid flow are strongly coupled, and capillary effects also introduce strong nonlinearity, the multi-objective optimization is very challenging. Despite this challenging setup, the PINN-THM results are in agreement with those reported by \citet{khoei2021modeling} and the subtle differences can be attributed to the phase-change effect, which is ignored by the PINN solver. 
The simulations show how the sudden change in the temperature and capillary pressure at the top surface develops through the domain until it reaches steady-state conditions, and how this results in the displacement of the porous layer.

\subsection{Injection-Production Within a Rigid Reservoir}
\begin{figure}[H]
    \centering
    \includegraphics[width=0.5\textwidth]{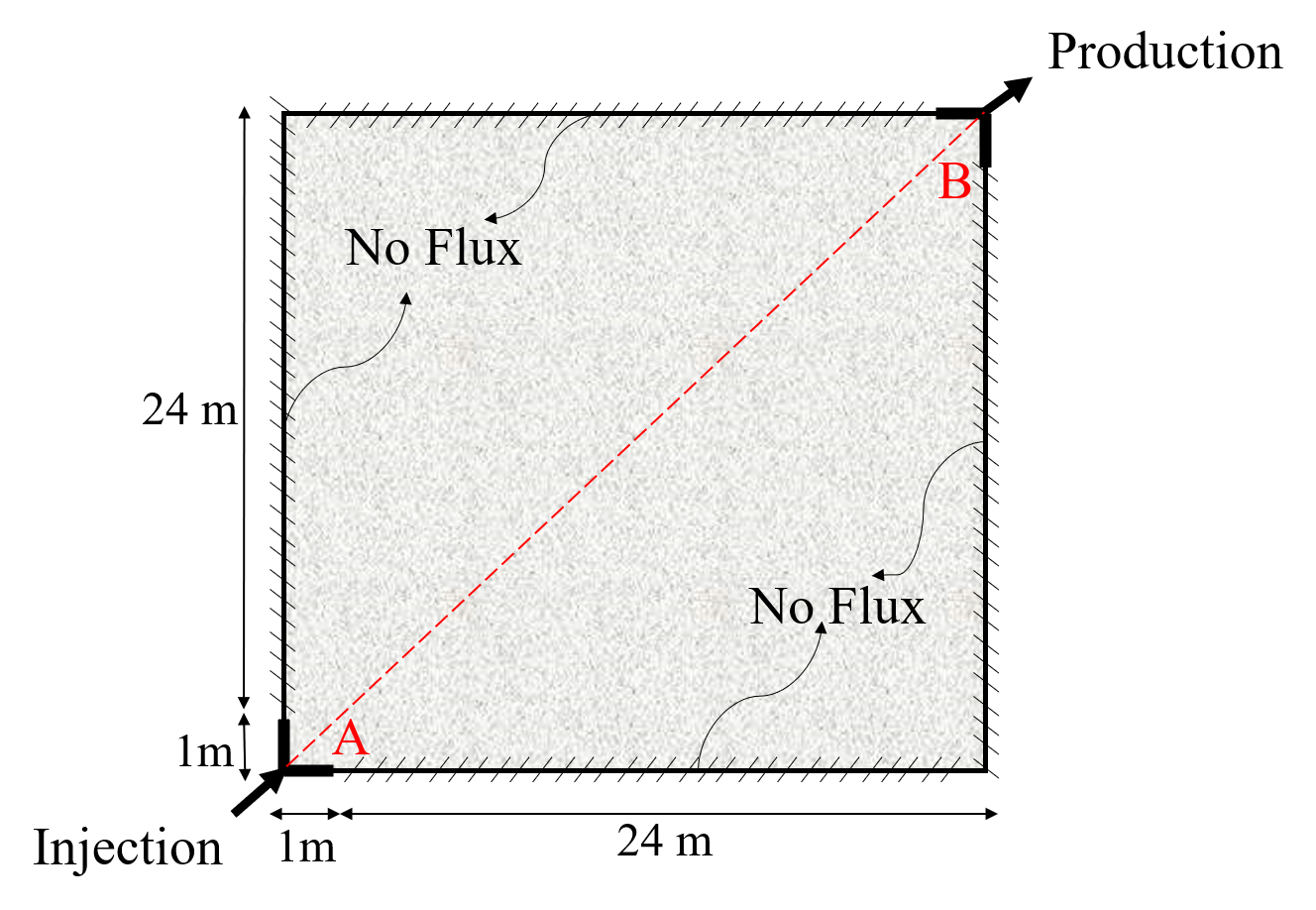}
    \caption{Geometry and boundary conditions of the injection-production in a rigid reservoir}
    \label{fig:geometry-injc-prod}
\end{figure}

As the last application example, let us consider the application of the proposed approach for fluid flow and heat transfer within a two-dimensional reservoir of dimensions $25~\text{m} \times 25~\text{m}$. The reservoir is assumed to be fully saturated, and the solid phase deformation is ignored, as considered by \citet{pao2001fully}. 
The initial pressure and temperature are set to $50~\text{MPa}$ and a reference value of $0~{^\circ}\text{C}$, respectively. Fluid injection takes place at the bottom-left corner, where a constant pressure equal to the initial pressure and a constant  temperature $75~{^\circ}\text{C}$ are imposed. Fluid production takes place at the top-right corner, where the  pressure is prescribed to a reference value of $0~\text{MPa}$. All other faces are considered impervious and thermally isolated, as shown in \Cref{fig:geometry-injc-prod}. 

The density and viscosity of water are taken as  ${\rho_w}=1000~\text{kg}/\text{m}^3$ and ${\mu_w}=0.5 \times 10^{-3}~\text{Pa} \cdot \text{s}$. The density of the solid is taken as ${\rho_s}=1800~\text{kg}/\text{m}^3$, and the porosity of the medium is taken as ${\phi}=0.1867$. Compressibility of the solid and fluid phases are ${c_s}=1.45 \times 10^{-8}~\text{1/Pa}$ and ${c_w}=4.35 \times 10^{-8}~\text{1/Pa}$, respectively. The thermal conductivity of the mixture is considered constant, as ${\lambda_\text{avg}}=5~\text{W}/\text{kg}\cdot{^\circ}\text{C}$.  Specific heat capacity of rock and water is assumed to be ${C_s}=200~\text{J}/\text{kg}\cdot{^\circ}\text{C}$ and ${C_w}=4184~\text{J}/\text{kg}\cdot{^\circ}\text{C}$. The intrinsic permeability and thermal expansion coefficient for rock and water are ${k}=0.09 \times 10^{-14}~\text{m}^2$, ${\beta_s}=9\times{10^{-7}}~{1}/{^\circ}\text{C}$, and ${\beta_w}=6.3\times{10^{-6}}~{1}/{^\circ}\text{C}$. In this example, the high pressure gradient between the injection and production areas accelerates the heat transfer through conduction.

\begin{figure}[H]
    \centering
    \includegraphics[width=1.0\textwidth]{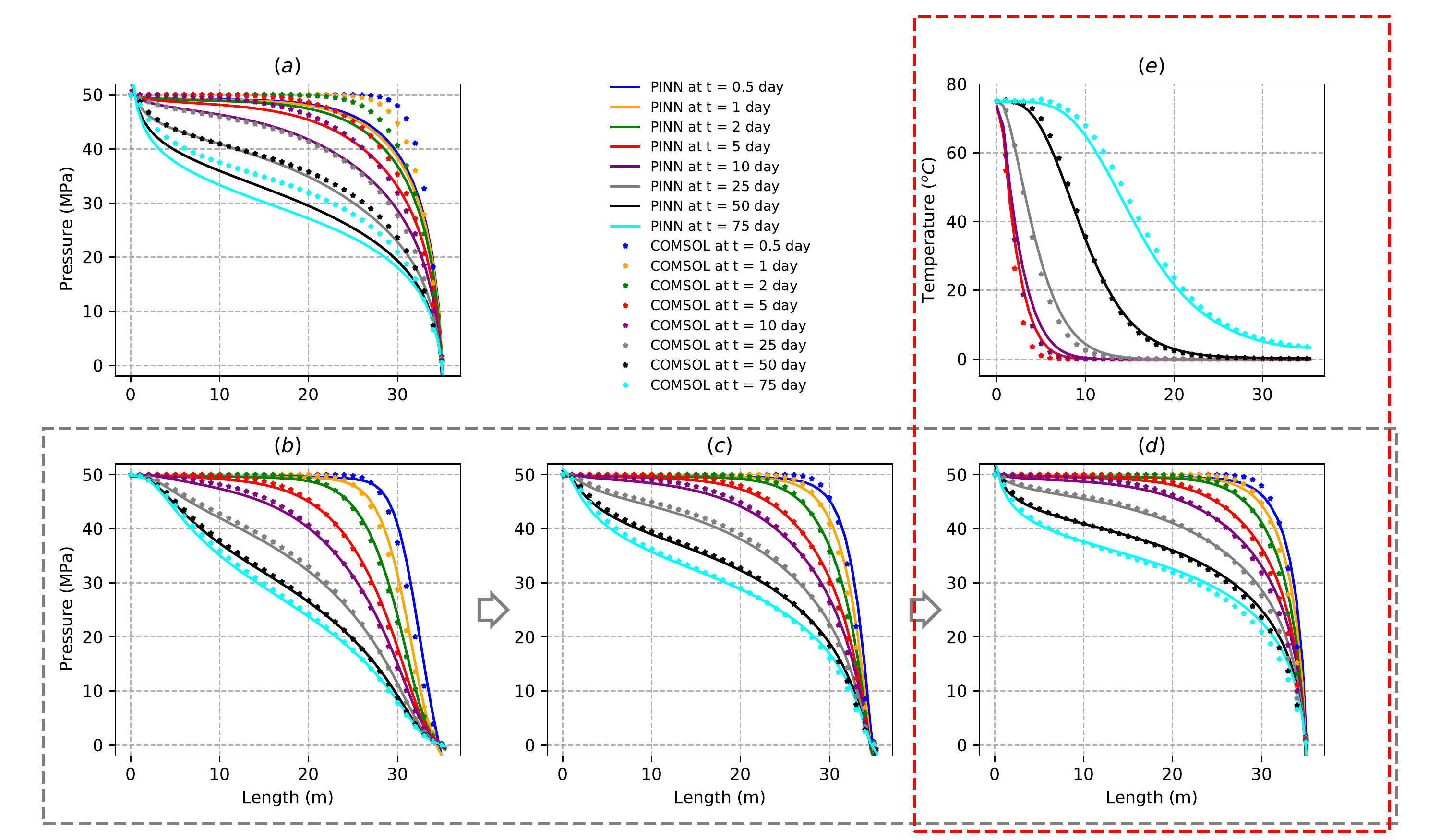}
    \caption{PINN solution (solid lines) along the domain diagonal (line A-B in \Cref{fig:geometry-injc-prod}) of the two-dimensional injection/production problem and comparison against reference results from COMSOL (symbols). Grey dashed box highlights the sequence of transfer learning. Red dashed box highlights the final PINN solutions for pressure and temperature along the diagonal. (a) PINN solution for pressure at different times without the use of transfer learning. (b-d) Pressure evolution by first solving an injection-production problem with 5m length (b), then transferring the weights to solve an injection-production problem with 2m length, and finally transferring the network parameters to solve the actual injection/production problem with 1m length. (e) Temperature evolution along the A-B diagonal line at different times, after transfer learning. }
    \label{fig:PT_all}
\end{figure}

\begin{figure}[H]
    \centering
    \includegraphics[width=1.0\textwidth]{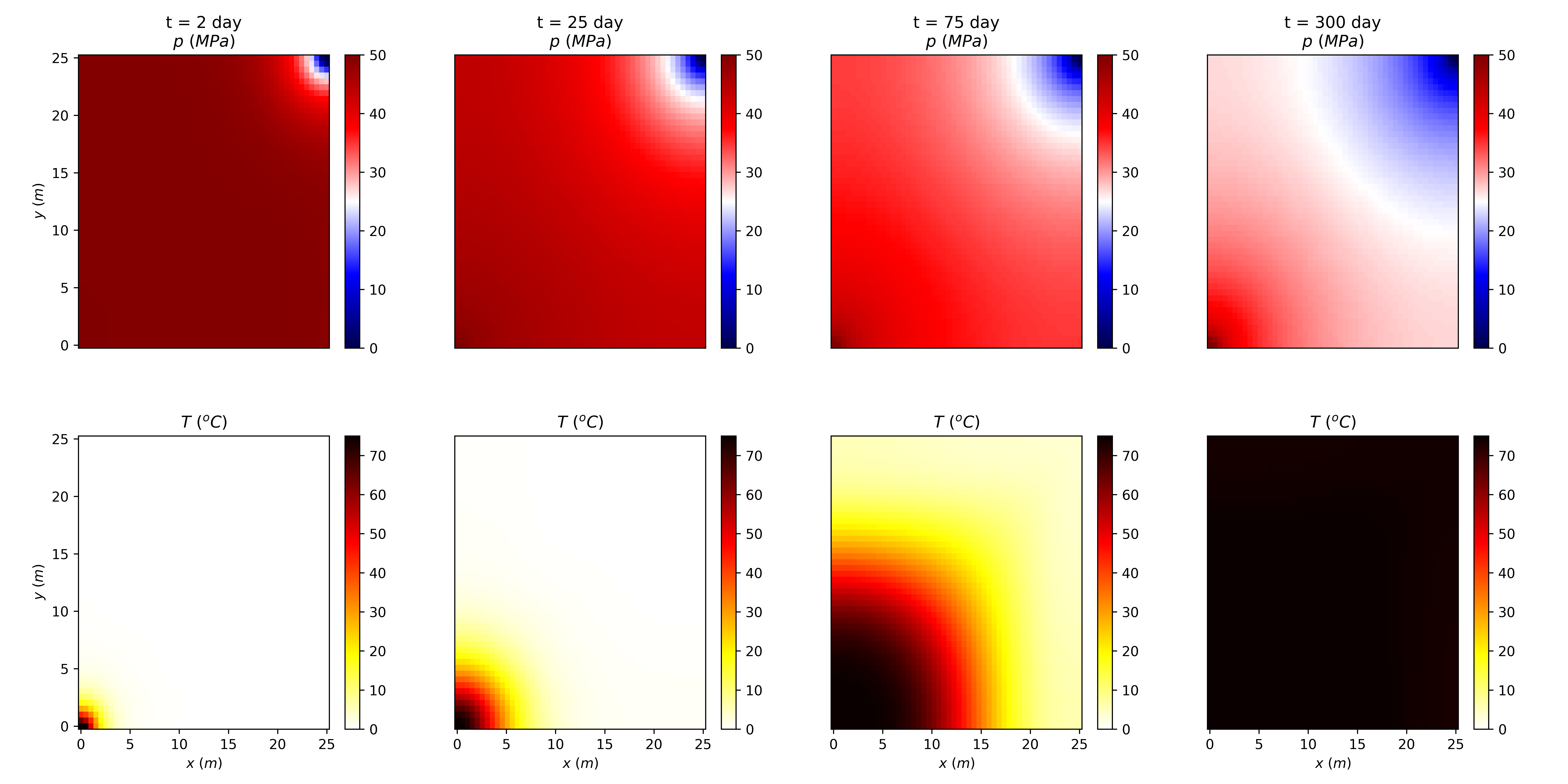}\caption{PINN's 2D plots for pressure and temperature over time for 1m injection/production length with transfer learning.}  \label{fig:PTcontour_1m}
\end{figure}

Due to the combination of boundary conditions, i.e., partly Neumann and partly Dirichlet, and the small injection and production lengths, we experienced that the PINN solver fails to find an accurate solution, regardless of the adaptive-weighting strategy used, as shown in \Cref{fig:PT_all}-a. 
We associate this behavior with the small injection-production regions and the sharp gradients in the vicinity of the transition point from Dirichlet- to Neumann-type boundary conditions. To resolve this, we used a remedy based on transfer learning. 
We first used a milder condition of picking an injection/production length of ${5}~\text{m}$, to pre-train the network parameters. We then transferred the trained parameters to an intermediate problem with ${2}~\text{m}$ injection-production lengths and finally back to the original one with ${1}~\text{m}$ length and re-performed the training. This transfer-learning progress is shown in \Cref{fig:PT_all}-b to d (inside the grey dashed box). The final results of pressure and temperature histories along the diagonal are plotted in \Cref{fig:PT_all}-d-e, which are in good agreement with the reference solution from COMSOL. The field plots for temperature and pressure distribution at different times are shown in \Cref{fig:PTcontour_1m}.

\section{Conclusions}
In this study, we have employed the PINN framework to simulate thermo-hydro-mechanical (THM) phenomena in porous media. Governing laws consist of linear momentum, mass conservation, and energy balance. Darcy's and Fourier's laws are employed to describe the fluid flow and heat transfer, respectively. 
Due to the complexity in optimizing the deep neural network for such a multiphysics system, we used a sequential training strategy. We applied the framework to several benchmark problems and reported good agreement with reference solutions. The combination of a sequential training strategy, a dimensionless formulation, adaptive weight methods, and transfer learning resulted in a good performance of PINN for simulating THM problems. 

While the PINN solver described above succeeded at simulating THM problems, we found that the network training was challenging, and slow compared with FEM solvers---suggesting that the training time remains a bottleneck of the framework. Thus, we believe that the PINNs might be better suited for solving inverse problems or for building surrogate models.

\bibliographystyle{abbrvnat} 
\bibliography{mybibfile}

\begin{thebibliography}{63}
\providecommand{\natexlab}[1]{#1}
\providecommand{\url}[1]{\texttt{#1}}
\expandafter\ifx\csname urlstyle\endcsname\relax
  \providecommand{\doi}[1]{doi: #1}\else
  \providecommand{\doi}{doi: \begingroup \urlstyle{rm}\Url}\fi

\bibitem[Abadi et~al.(2016)Abadi, Barham, Chen, Chen, Davis, Dean, Devin,
  Ghemawat, Irving, Isard, et~al.]{abadi2016tensorflow}
M.~Abadi, P.~Barham, J.~Chen, Z.~Chen, A.~Davis, J.~Dean, M.~Devin,
  S.~Ghemawat, G.~Irving, M.~Isard, et~al.
\newblock {TensorFlow}: A system for large-scale machine learning.
\newblock In \emph{12th USENIX symposium on operating systems design and
  implementation (OSDI 16)}, pages 265--283, 2016.

\bibitem[Almajid and Abu-Alsaud(2021)]{43-PINN-similar-Hamdi}
M.~M. Almajid and M.~O. Abu-Alsaud.
\newblock Prediction of porous media fluid flow using physics informed neural
  networks.
\newblock \emph{Journal of Petroleum Science and Engineering}, 208:\penalty0
  109205, 2021.

\bibitem[Armero and Simo(1992)]{armero1992new}
F.~Armero and J.~Simo.
\newblock A new unconditionally stable fractional step method for non-linear
  coupled thermomechanical problems.
\newblock \emph{International Journal for Numerical Methods in Engineering},
  35\penalty0 (4):\penalty0 737--766, 1992.

\bibitem[Ballarini et~al.(2017)Ballarini, Graupner, and
  Bauer]{ballarini2017thermal}
E.~Ballarini, B.~Graupner, and S.~Bauer.
\newblock Thermal--hydraulic--mechanical behavior of bentonite and
  sand-bentonite materials as seal for a nuclear waste repository: Numerical
  simulation of column experiments.
\newblock \emph{Applied Clay Science}, 135:\penalty0 289--299, 2017.

\bibitem[Baydin et~al.(2018)Baydin, Pearlmutter, Radul, and
  Siskind]{baydin2018automatic}
A.~G. Baydin, B.~A. Pearlmutter, A.~A. Radul, and J.~M. Siskind.
\newblock Automatic differentiation in machine learning: a survey.
\newblock \emph{Journal of Machine Learning Research}, 18:\penalty0 1--23,
  2018.

\bibitem[Bekele(2020)]{36-PINN-BarryMercer}
Y.~W. Bekele.
\newblock Physics-informed deep learning for flow and deformation in
  poroelastic media.
\newblock \emph{arXiv preprint arXiv:2010.15426}, 2020.

\bibitem[Bekele(2021)]{41-PINNTerzaghi-forward-inverse}
Y.~W. Bekele.
\newblock Physics-informed deep learning for one-dimensional consolidation.
\newblock \emph{Journal of Rock Mechanics and Geotechnical Engineering},
  13\penalty0 (2):\penalty0 420--430, 2021.

\bibitem[Brooks and Corey(1966)]{Brooks-Corey}
R.~H. Brooks and A.~T. Corey.
\newblock Properties of porous media affecting fluid flow.
\newblock \emph{Journal of the Irrigation and Drainage Division}, 92\penalty0
  (2):\penalty0 61--88, 1966.

\bibitem[Cai et~al.(2021)Cai, Wang, Wang, Perdikaris, and
  Karniadakis]{cai2021physicsheat}
S.~Cai, Z.~Wang, S.~Wang, P.~Perdikaris, and G.~E. Karniadakis.
\newblock Physics-informed neural networks for heat transfer problems.
\newblock \emph{Journal of Heat Transfer}, 143\penalty0 (6):\penalty0 060801,
  2021.

\bibitem[Cai et~al.(2022)Cai, Mao, Wang, Yin, and Karniadakis]{cai2021physics}
S.~Cai, Z.~Mao, Z.~Wang, M.~Yin, and G.~E. Karniadakis.
\newblock Physics-informed neural networks ({PINNs}) for fluid mechanics: A
  review.
\newblock \emph{Acta Mechanica Sinica}, pages 1--12, 2022.

\bibitem[Coussy(2004)]{coussy2004poromechanics}
O.~Coussy.
\newblock \emph{Poromechanics}.
\newblock John Wiley \& Sons, 2004.

\bibitem[Dakshanamurthy and Fredlund(1981)]{14-THM-1981}
V.~Dakshanamurthy and D.~Fredlund.
\newblock A mathematical model for predicting moisture flow in an unsaturated
  soil under hydraulic and temperature gradients.
\newblock \emph{Water Resources Research}, 17\penalty0 (3):\penalty0 714--722,
  1981.

\bibitem[Eivazi et~al.(2021)Eivazi, Tahani, Schlatter, and
  Vinuesa]{eivazi2021physics}
H.~Eivazi, M.~Tahani, P.~Schlatter, and R.~Vinuesa.
\newblock Physics-informed neural networks for solving {Reynolds}-averaged
  {Navier Stokes} equations.
\newblock \emph{arXiv preprint arXiv:2107.10711}, 2021.

\bibitem[Fraces and Tchelepi(2021)]{16-PINN-Hamdi-2D-2phases}
C.~G. Fraces and H.~Tchelepi.
\newblock Physics informed deep learning for flow and transport in porous
  media.
\newblock In \emph{SPE Reservoir Simulation Conference}. OnePetro, 2021.

\bibitem[Fraces et~al.(2020)Fraces, Papaioannou, and
  Tchelepi]{14-PINN-Hamdi-GAN-inversion}
C.~G. Fraces, A.~Papaioannou, and H.~Tchelepi.
\newblock Physics informed deep learning for transport in porous media. buckley
  leverett problem.
\newblock \emph{arXiv preprint arXiv:2001.05172}, 2020.

\bibitem[Fuks and Tchelepi(2020)]{15-PINN-Hamdi-limitations}
O.~Fuks and H.~A. Tchelepi.
\newblock Limitations of physics informed machine learning for nonlinear
  two-phase transport in porous media.
\newblock \emph{Journal of Machine Learning for Modeling and Computing},
  1\penalty0 (1), 2020.

\bibitem[Garipov et~al.(2018)Garipov, Tomin, Rin, Voskov, and
  Tchelepi]{15-THM-2018}
T.~T. Garipov, P.~Tomin, R.~Rin, D.~V. Voskov, and H.~A. Tchelepi.
\newblock Unified thermo-compositional-mechanical framework for reservoir
  simulation.
\newblock \emph{Computational Geosciences}, 22\penalty0 (4):\penalty0
  1039--1057, 2018.

\bibitem[Gawin et~al.(1996)Gawin, Schrefler, and Galindo]{8-THM-1996}
D.~Gawin, B.~A. Schrefler, and M.~Galindo.
\newblock Thermo-hydro-mechanical analysis of partially saturated porous
  materials.
\newblock \emph{Engineering Computations}, 13\penalty0 (7):\penalty0 113--143,
  1996.

\bibitem[Gelet et~al.(2012)Gelet, Loret, and Khalili]{gelet2012thermo}
R.~Gelet, B.~Loret, and N.~Khalili.
\newblock A thermo-hydro-mechanical coupled model in local thermal
  non-equilibrium for fractured hdr reservoir with double porosity.
\newblock \emph{Journal of Geophysical Research: Solid Earth}, 117\penalty0
  (B7), 2012.

\bibitem[Goodfellow et~al.(2016)Goodfellow, Bengio, and
  Courville]{goodfellow2016deep}
I.~Goodfellow, Y.~Bengio, and A.~Courville.
\newblock \emph{Deep Learning}.
\newblock MIT press, 2016.

\bibitem[Haghighat and Juanes(2021)]{haghighat2021sciann}
E.~Haghighat and R.~Juanes.
\newblock {SciANN}: A {Keras}/{TensorFlow} wrapper for scientific computations
  and physics-informed deep learning using artificial neural networks.
\newblock \emph{Computer Methods in Applied Mechanics and Engineering},
  373:\penalty0 113552, 2021.

\bibitem[Haghighat et~al.(2021{\natexlab{a}})Haghighat, Amini, and
  Juanes]{our-HM-paper}
E.~Haghighat, D.~Amini, and R.~Juanes.
\newblock Physics-informed neural network simulation of multiphase
  poroelasticity using stress-split sequential training.
\newblock \emph{arXiv preprint arXiv:2110.03049}, 2021{\natexlab{a}}.

\bibitem[Haghighat et~al.(2021{\natexlab{b}})Haghighat, Bekar, Madenci, and
  Juanes]{haghighat2021nonlocal}
E.~Haghighat, A.~C. Bekar, E.~Madenci, and R.~Juanes.
\newblock A nonlocal physics-informed deep learning framework using the
  peridynamic differential operator.
\newblock \emph{Computer Methods in Applied Mechanics and Engineering},
  385:\penalty0 114012, 2021{\natexlab{b}}.

\bibitem[Haghighat et~al.(2021{\natexlab{c}})Haghighat, Raissi, Moure, Gomez,
  and Juanes]{haghighat2021physics}
E.~Haghighat, M.~Raissi, A.~Moure, H.~Gomez, and R.~Juanes.
\newblock A physics-informed deep learning framework for inversion and
  surrogate modeling in solid mechanics.
\newblock \emph{Computer Methods in Applied Mechanics and Engineering},
  379:\penalty0 113741, 2021{\natexlab{c}}.

\bibitem[Hughes(2012)]{hughes2012finite}
T.~J. Hughes.
\newblock \emph{The Finite Element Method: Linear Static and Dynamic Finite
  Element Analysis}.
\newblock Courier Corporation, 2012.

\bibitem[Iranmanesh et~al.(2018)Iranmanesh, Pak, and Samimi]{20-THM-pak}
M.~A. Iranmanesh, A.~Pak, and S.~Samimi.
\newblock Non-isothermal simulation of the behavior of unsaturated soils using
  a novel efg-based three dimensional model.
\newblock \emph{Computers and Geotechnics}, 99:\penalty0 93--103, 2018.

\bibitem[Jagtap and Karniadakis(2020)]{33-PINN-XPINN}
A.~D. Jagtap and G.~E. Karniadakis.
\newblock Extended physics-informed neural networks ({XPINNs}): A generalized
  space-time domain decomposition based deep learning framework for nonlinear
  partial differential equations.
\newblock \emph{Communications in Computational Physics}, 28\penalty0
  (5):\penalty0 2002--2041, 2020.

\bibitem[Jagtap et~al.(2020)Jagtap, Kawaguchi, and
  Karniadakis]{1-PINN-AddaptiveActivationFunction}
A.~D. Jagtap, K.~Kawaguchi, and G.~E. Karniadakis.
\newblock Adaptive activation functions accelerate convergence in deep and
  physics-informed neural networks.
\newblock \emph{Journal of Computational Physics}, 404:\penalty0 109136, 2020.

\bibitem[Jha and Juanes(2014)]{jhajuanes14-wrr}
B.~Jha and R.~Juanes.
\newblock Coupled multiphase flow and poromechanics: a computational model of
  pore-pressure effects on fault slip and earthquake triggering.
\newblock \emph{Water Resources Research}, 50\penalty0 (5):\penalty0
  3776--3808, doi:10.1002/2013WR015175, 2014.

\bibitem[Jin et~al.(2021)Jin, Cai, Li, and Karniadakis]{jin2021nsfnets}
X.~Jin, S.~Cai, H.~Li, and G.~E. Karniadakis.
\newblock {NSFnets} ({Navier-Stokes} flow nets): Physics-informed neural
  networks for the incompressible {Navier-Stokes} equations.
\newblock \emph{Journal of Computational Physics}, 426:\penalty0 109951, 2021.

\bibitem[Kadeethum et~al.(2020)Kadeethum, J{\o}rgensen, and
  Nick]{24-PINN-Biot-forward-inversion}
T.~Kadeethum, T.~M. J{\o}rgensen, and H.~M. Nick.
\newblock Physics-informed neural networks for solving nonlinear diffusivity
  and biot’s equations.
\newblock \emph{PloS One}, 15\penalty0 (5):\penalty0 e0232683, 2020.

\bibitem[Karniadakis et~al.(2021)Karniadakis, Kevrekidis, Lu, Perdikaris, Wang,
  and Yang]{39-PINN-TotalReview}
G.~E. Karniadakis, I.~G. Kevrekidis, L.~Lu, P.~Perdikaris, S.~Wang, and
  L.~Yang.
\newblock Physics-informed machine learning.
\newblock \emph{Nature Reviews Physics}, 3\penalty0 (6):\penalty0 422--440,
  2021.

\bibitem[Khoei et~al.(2021)Khoei, Amini, and Mortazavi]{khoei2021modeling}
A.~R. Khoei, D.~Amini, and S.~M.~S. Mortazavi.
\newblock Modeling non-isothermal two-phase fluid flow with phase change in
  deformable fractured porous media using extended finite element method.
\newblock \emph{International Journal for Numerical Methods in Engineering},
  122:\penalty0 4378--4426, 2021.

\bibitem[Kim(2018)]{kim2018unconditionally}
J.~Kim.
\newblock Unconditionally stable sequential schemes for all-way coupled
  thermoporomechanics: Undrained-adiabatic and extended fixed-stress splits.
\newblock \emph{Computer Methods in Applied Mechanics and Engineering},
  341:\penalty0 93--112, 2018.

\bibitem[Kim et~al.(2011{\natexlab{a}})Kim, Tchelepi, and
  Juanes]{kim-fixedStress}
J.~Kim, H.~A. Tchelepi, and R.~Juanes.
\newblock Stability and convergence of sequential methods for coupled flow and
  geomechanics: Fixed-stress and fixed-strain splits.
\newblock \emph{Computer Methods in Applied Mechanics and Engineering},
  200\penalty0 (13-16):\penalty0 1591--1606, 2011{\natexlab{a}}.

\bibitem[Kim et~al.(2011{\natexlab{b}})Kim, Tchelepi, and
  Juanes]{kimtchelepijuanes11-drained}
J.~Kim, H.~A. Tchelepi, and R.~Juanes.
\newblock Stability and convergence of sequential methods for coupled flow and
  geomechanics: {D}rained and undrained splits.
\newblock \emph{Computer Methods in Applied Mechanics and Engineering},
  200:\penalty0 2094--2116, 2011{\natexlab{b}}.

\bibitem[Kim et~al.(2013)Kim, Tchelepi, and Juanes]{kim2013rigorous}
J.~Kim, H.~A. Tchelepi, and R.~Juanes.
\newblock Rigorous coupling of geomechanics and multiphase flow with strong
  capillarity.
\newblock \emph{SPE Journal}, 18\penalty0 (06):\penalty0 1123--1139, 2013.

\bibitem[Kingma and Ba(2014)]{kingma2014adam}
D.~P. Kingma and J.~Ba.
\newblock Adam: A method for stochastic optimization.
\newblock \emph{arXiv preprint arXiv:1412.6980}, 2014.

\bibitem[Laubscher(2021)]{23-PINN-mulit-species-flow-heat-sequential}
R.~Laubscher.
\newblock Simulation of multi-species flow and heat transfer using
  physics-informed neural networks.
\newblock \emph{Physics of Fluids}, 33\penalty0 (8):\penalty0 087101, 2021.

\bibitem[Lewis et~al.(1998)Lewis, Lewis, and Schrefler]{Lewis-book}
R.~W. Lewis, R.~W. Lewis, and B.~Schrefler.
\newblock \emph{The Finite Element Method in the Static and Dynamic Deformation
  and Consolidation of Porous Media}.
\newblock John Wiley \& Sons, 1998.

\bibitem[Mao et~al.(2020)Mao, Jagtap, and Karniadakis]{mao2020physics}
Z.~Mao, A.~D. Jagtap, and G.~E. Karniadakis.
\newblock Physics-informed neural networks for high-speed flows.
\newblock \emph{Computer Methods in Applied Mechanics and Engineering},
  360:\penalty0 112789, 2020.

\bibitem[Niaki et~al.(2021)Niaki, Haghighat, Campbell, Poursartip, and
  Vaziri]{niaki2021physics}
S.~A. Niaki, E.~Haghighat, T.~Campbell, A.~Poursartip, and R.~Vaziri.
\newblock Physics-informed neural network for modelling the thermochemical
  curing process of composite-tool systems during manufacture.
\newblock \emph{Computer Methods in Applied Mechanics and Engineering},
  384:\penalty0 113959, 2021.

\bibitem[Nocedal and Wright(2006)]{nocedal2006numerical}
J.~Nocedal and S.~Wright.
\newblock \emph{Numerical Optimization}.
\newblock Springer Science \& Business Media, 2006.

\bibitem[Noorishad et~al.(1984)Noorishad, Tsang, and Witherspoon]{3-THM-1984}
J.~Noorishad, C.~Tsang, and P.~Witherspoon.
\newblock Coupled thermal-hydraulic-mechanical phenomena in saturated fractured
  porous rocks: Numerical approach.
\newblock \emph{Journal of Geophysical Research: Solid Earth}, 89\penalty0
  (B12):\penalty0 10365--10373, 1984.

\bibitem[O'Sullivan et~al.(2001)O'Sullivan, Pruess, and Lippmann]{o2001state}
M.~J. O'Sullivan, K.~Pruess, and M.~J. Lippmann.
\newblock State of the art of geothermal reservoir simulation.
\newblock \emph{Geothermics}, 30\penalty0 (4):\penalty0 395--429, 2001.

\bibitem[Pandey et~al.(2017)Pandey, Chaudhuri, and Kelkar]{pandey2017coupled}
S.~Pandey, A.~Chaudhuri, and S.~Kelkar.
\newblock A coupled thermo-hydro-mechanical modeling of fracture aperture
  alteration and reservoir deformation during heat extraction from a geothermal
  reservoir.
\newblock \emph{Geothermics}, 65:\penalty0 17--31, 2017.

\bibitem[Pao et~al.(2001)Pao, Lewis, and Masters]{pao2001fully}
W.~K. Pao, R.~W. Lewis, and I.~Masters.
\newblock A fully coupled hydro-thermo-poro-mechanical model for black oil
  reservoir simulation.
\newblock \emph{International Journal for Numerical and Analytical Methods in
  Geomechanics}, 25\penalty0 (12):\penalty0 1229--1256, 2001.

\bibitem[Praditia et~al.(2018)Praditia, Helmig, and
  Hajibeygi]{praditia2018multiscale}
T.~Praditia, R.~Helmig, and H.~Hajibeygi.
\newblock Multiscale formulation for coupled flow-heat equations arising from
  single-phase flow in fractured geothermal reservoirs.
\newblock \emph{Computational Geosciences}, 22\penalty0 (5):\penalty0
  1305--1322, 2018.

\bibitem[Raissi et~al.(2019)Raissi, Perdikaris, and
  Karniadakis]{IntroductionPINN}
M.~Raissi, P.~Perdikaris, and G.~E. Karniadakis.
\newblock Physics-informed neural networks: A deep learning framework for
  solving forward and inverse problems involving nonlinear partial differential
  equations.
\newblock \emph{Journal of Computational Physics}, 378:\penalty0 686--707,
  2019.

\bibitem[Ranade et~al.(2021)Ranade, Hill, and
  Pathak]{ranade2021discretizationnet}
R.~Ranade, C.~Hill, and J.~Pathak.
\newblock {DiscretizationNet}: A machine-learning based solver for
  {Navier--Stokes} equations using finite volume discretization.
\newblock \emph{Computer Methods in Applied Mechanics and Engineering},
  378:\penalty0 113722, 2021.

\bibitem[Rao et~al.(2020)Rao, Sun, and Liu]{rao2020physics}
C.~Rao, H.~Sun, and Y.~Liu.
\newblock Physics-informed deep learning for incompressible laminar flows.
\newblock \emph{Theoretical and Applied Mechanics Letters}, 10\penalty0
  (3):\penalty0 207--212, 2020.

\bibitem[Rao et~al.(2021)Rao, Sun, and Liu]{rao2021physics}
C.~Rao, H.~Sun, and Y.~Liu.
\newblock Physics-informed deep learning for computational elastodynamics
  without labeled data.
\newblock \emph{Journal of Engineering Mechanics}, 147\penalty0 (8):\penalty0
  04021043, 2021.

\bibitem[Reyes et~al.(2021)Reyes, Howard, Perdikaris, and
  Tartakovsky]{reyes2021learning}
B.~Reyes, A.~A. Howard, P.~Perdikaris, and A.~M. Tartakovsky.
\newblock Learning unknown physics of non-{Newtonian} fluids.
\newblock \emph{Physical Review Fluids}, 6\penalty0 (7):\penalty0 073301, 2021.

\bibitem[Rutqvist et~al.(2001)Rutqvist, B{\"o}rgesson, Chijimatsu, Kobayashi,
  Jing, Nguyen, Noorishad, and Tsang]{rutqvist2001thermohydromechanics}
J.~Rutqvist, L.~B{\"o}rgesson, M.~Chijimatsu, A.~Kobayashi, L.~Jing, T.~Nguyen,
  J.~Noorishad, and C.-F. Tsang.
\newblock Thermohydromechanics of partially saturated geological media:
  governing equations and formulation of four finite element models.
\newblock \emph{International Journal of Rock Mechanics and Mining Sciences},
  38\penalty0 (1):\penalty0 105--127, 2001.

\bibitem[Rutqvist et~al.(2009)Rutqvist, Barr, Birkholzer, Fujisaki, Kolditz,
  Liu, Fujita, Wang, and Zhang]{rutqvist2009comparative}
J.~Rutqvist, D.~Barr, J.~T. Birkholzer, K.~Fujisaki, O.~Kolditz, Q.-S. Liu,
  T.~Fujita, W.~Wang, and C.-Y. Zhang.
\newblock A comparative simulation study of coupled {THM} processes and their
  effect on fractured rock permeability around nuclear waste repositories.
\newblock \emph{Environmental Geology}, 57\penalty0 (6):\penalty0 1347--1360,
  2009.

\bibitem[Schrefler et~al.(1997)Schrefler, Simoni, and Turska]{23-THM}
B.~Schrefler, L.~Simoni, and E.~Turska.
\newblock Standard staggered and staggered newton schemes in
  thermo-hydro-mechanical problems.
\newblock \emph{Computer Methods in Applied Mechanics and Engineering},
  144\penalty0 (1-2):\penalty0 93--109, 1997.

\bibitem[Schrefler et~al.(1995)Schrefler, Zhan, and Simoni]{13-THM}
B.~A. Schrefler, X.~Zhan, and L.~Simoni.
\newblock A coupled model for water flow, airflow and heat flow in deformable
  porous media.
\newblock \emph{International Journal of Numerical Methods for Heat \& Fluid
  Flow}, 5\penalty0 (6):\penalty0 531--547, 1995.

\bibitem[Wang et~al.(2021)Wang, Wang, and Perdikaris]{10-PINN-FourierNetwork}
S.~Wang, H.~Wang, and P.~Perdikaris.
\newblock On the eigenvector bias of {Fourier} feature networks: From
  regression to solving multi-scale {PDEs} with physics-informed neural
  networks.
\newblock \emph{Computer Methods in Applied Mechanics and Engineering},
  384:\penalty0 113938, 2021.

\bibitem[Wang et~al.(2022)Wang, Yu, and Perdikaris]{wang2022and}
S.~Wang, X.~Yu, and P.~Perdikaris.
\newblock When and why {PINNs} fail to train: A neural tangent kernel
  perspective.
\newblock \emph{Journal of Computational Physics}, 449:\penalty0 110768, 2022.

\bibitem[Wang et~al.(2009)Wang, Kosakowski, and Kolditz]{wang2009parallel}
W.~Wang, G.~Kosakowski, and O.~Kolditz.
\newblock A parallel finite element scheme for thermo-hydro-mechanical ({THM})
  coupled problems in porous media.
\newblock \emph{Computers \& Geosciences}, 35\penalty0 (8):\penalty0
  1631--1641, 2009.

\bibitem[Wu et~al.(2018)Wu, Xiao, and Paterson]{wu2018physics}
J.-L. Wu, H.~Xiao, and E.~Paterson.
\newblock Physics-informed machine learning approach for augmenting turbulence
  models: A comprehensive framework.
\newblock \emph{Physical Review Fluids}, 3\penalty0 (7):\penalty0 074602, 2018.

\bibitem[Yu et~al.(2018)]{yu2017deep}
B.~Yu et~al.
\newblock The deep {Ritz} method: a deep learning-based numerical algorithm for
  solving variational problems.
\newblock \emph{Communications in Mathematics and Statistics}, 6\penalty0
  (1):\penalty0 1--12, 2018.

\bibitem[Zienkiewicz et~al.(1977)Zienkiewicz, Taylor, Nithiarasu, and
  Zhu]{zienkiewicz1977finite}
O.~C. Zienkiewicz, R.~L. Taylor, P.~Nithiarasu, and J.~Zhu.
\newblock \emph{The Finite Element Method}, volume~3.
\newblock McGraw-hill London, 1977.

\end{thebibliography}

\end{document}